\documentclass[letterpaper]{article} % DO NOT CHANGE THIS
\usepackage[preprint]{aaai2027} % Camera-Ready Use
\usepackage[hyphens]{url}  % DO NOT CHANGE THIS
\usepackage{graphicx} % DOarticle NOT CHANGE THIS
\usepackage{natbib}  % DO NOT CHANGE THIS AND DO NOT ADD ANY OPTIONS TO IT
\usepackage{caption} % DO NOT CHANGE THIS AND DO NOT ADD ANY OPTIONS TO IT
\usepackage{algorithm}
\usepackage{algorithmic}
\usepackage{amsmath}
\usepackage{xcolor}
\usepackage[table]{xcolor}
\usepackage{booktabs}
\usepackage{multirow}
\usepackage{amssymb}
\usepackage{bbding}
\usepackage{subfig}
\usepackage{pifont}
\usepackage{makecell}
\definecolor{ggray}{RGB}{120,120,120}

\definecolor{bestbg}{RGB}{198,234,212}   % 柔和的绿色（色盲友好）
\definecolor{secondbg}{RGB}{231,242,255} % 淡蓝，给次优
\definecolor{secondbg2}{RGB}{255, 218, 224}

\newcommand{\boldparagraph}[1]{\noindent\textbf{#1}\ }

\newcommand{\methodname}{EffectLearner}
\newcommand{\datasetname}{EffectWorld}
\newcommand{\ourbencheval}{EffectWorld-Eval}
\newcommand{\ourbenchwild}{EffectWorld-Wild}
\newcommand{\suppl}{{{\textit{Supp. Mat.}}}}
\usepackage{fontawesome6} % 导言区加载

\usepackage[colorlinks=true,
linkcolor=blue,       % \\ref 交叉引用颜色
citecolor=blue,       % \\cite 文献引用颜色
urlcolor=blue,        % 网址颜色
pdfborder={0 0 0}     % 去除链接外框
]{hyperref}

\usepackage{newfloat}
\usepackage{listings}
\DeclareCaptionStyle{ruled}{labelfont=normalfont,labelsep=colon,strut=off} % DO NOT CHANGE THIS
\floatstyle{ruled}
\newfloat{listing}{tb}{lst}{}
\floatname{listing}{Listing}

\usepackage{booktabs}

\title{EffectLearner: World-Aware Object-Effect Reasoning for Real-World\\Video Object Removal}

\author {
    Feier Wu\textsuperscript{\rm 1}\equalcontrib,
    Wanke Xia\textsuperscript{\rm 1}\equalcontrib,
    Xu He\textsuperscript{\rm 1}\equalcontrib,
    Zilang Zhou\textsuperscript{\rm 2},
    Si Chen\textsuperscript{\rm 3},
    Dongxia Liu\textsuperscript{\rm 1},
    Liyang Chen\textsuperscript{\rm 1}, \\
    Qimeng Wu\textsuperscript{\rm 1},
    Zhengbo Zhang\textsuperscript{\rm 4},
    Wenming Yang\textsuperscript{\rm 1}\corresponding,
    Zhiyong Wu\textsuperscript{\rm 1}\corresponding
}
\affiliations {
    \textsuperscript{\rm 1}Tsinghua University,
    \textsuperscript{\rm 2}China Agricultural University, \\
    \textsuperscript{\rm 3}Jiangnan University,
    \textsuperscript{\rm 4}Institute of Automation, Chinese Academy of Science
    \\
    Project Site: \includegraphics[height=1em]{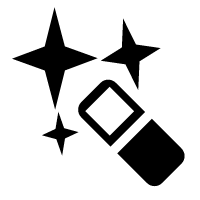}\;\url{https://morleyolsen.github.io/EffectLearner/}\\
    Dataset: \includegraphics[height=1em]{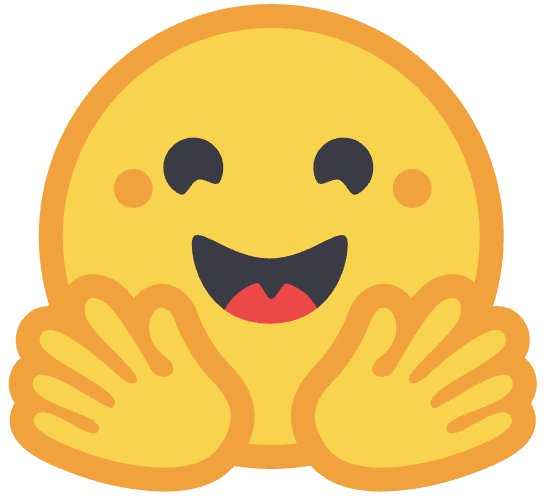}\;\url{https://huggingface.co/datasets/jenniferwuuu/EffectWorld}
}

\begin{document}

\maketitle
\begin{abstract}
Video object removal must eliminate not only the target object but also its induced effects while maintaining high-fidelity and spatiotemporally coherent restoration.
Existing methods mainly learn object-effect correspondences implicitly from predefined effect categories and fixed data distributions, limiting their generalization to complex real-world scenes involving compositional effects, spatially detached or weakly correlated effects, long-tail physical phenomena, and dynamically evolving interactions.
We propose \textbf{\methodname{}}, a semantic-reasoning-enhanced framework that combines a VLM-based Object-Effect Reasoner with a DiT-based Video Eraser.
Guided by a structured effect-analysis prompt, the Reasoner performs cross-modal reasoning over a target-highlighted video and extracts compact effect-aware context, which guides the Video Eraser toward comprehensive object-effect removal.
Motion-aware mask guidance and motion-consistency supervision further improve removal coverage and spatiotemporal stability under object motion and evolving scene dynamics.
To fully exploit the framework in challenging real-world scenarios, we further construct \textbf{\datasetname{}}, a paired video dataset specifically designed for complex object-induced effects, and introduce a progressive training curriculum that combines common supervision with complex-effect data.
On the standard ROSE-Bench, \methodname{} outperforms existing baselines on most metrics and achieves clear advantages on both \ourbencheval{} and the challenging \ourbenchwild{}, demonstrating its ability to deliver high-quality video object removal in complex real-world scenes.
\end{abstract}
\section{Introduction}
\label{intro}
%% P1 定义、应用、基本要求
% 视频物体移除旨在从视频中删除不需要的动态视觉内容，同时保持逼真的视觉外观和连贯的时间动态。
Video object removal (VOR) aims to eliminate unwanted dynamic content from videos while maintaining high visual fidelity and spatiotemporal coherence.
% 该技术正日益广泛应用于日常媒体编辑、数字内容创作和专业影视后期制作。
It is increasingly used in everyday media editing, digital content creation, and professional post-production~\cite{kushwaha2026object,huang2023recent,yu2024barriers}.
% 然而，高质量视频物体移除的核心挑战并不只是删除目标物体本身，还需要同时消除其引起的阴影、反射、光照变化、形变和动态痕迹。
High-quality VOR requires more than erasing the target itself, as it must also eliminate object-induced effects such as shadows, reflections, and illuminations.
In real-world scenarios, these effects often extend beyond the target region and evolve with object motion and scene dynamics, requiring both complete effect removal and spatiotemporally consistent restoration across frames.

%% P2 现有方法的缺点
Early inpainting-based object removal approaches typically rely on artificial pairs constructed by zero-masking or copy-pasting objects~\cite{chang2019vornet}.
Such supervision often leads models to remove only the target itself, leaving object-induced effects behind and producing unnatural results.
Building on advances in large-scale Diffusion Transformer (DiT)-based video generation~\cite{kong2024hunyuanvideo,peebles2023scalable}, recent efforts learn from physically grounded video pairs obtained through physics-based 3D rendering~\cite{zhang2024physdreamer,huang2025dreamphysics} or controlled real-world capture~\cite{li2025realcam,liu2026mozoo,mao2026omni,zhang2025egolcd}, enabling the joint removal of objects and their associated effects.
However, these methods still learn object-effect correspondences \emph{implicitly} from a limited set of predefined categories and fixed training distributions, without a dedicated mechanism to reason about how the target object interacts with its surroundings.
They therefore generalize poorly to complex real-world scenes involving compositional effects, weak object-effect correlations, and long-tail physical phenomena.
Examples include illumination changes spatially detached from their source, water ripples and wakes, smoke or dust, and persistent motion trails.
% ~\TODO{teaser? case}
%
Moreover, existing removal guidance and training objectives place greater emphasis on frame-wise erasure quality than on the spatiotemporal consistency of restored regions.
Consequently, when object motion and scene dynamics give rise to spatiotemporally evolving effects, these methods often produce inconsistent removal across frames, including missed removals, residual traces, and temporal flicker.
We argue that addressing these challenges requires reasoning about how objects interact with scenes and what effects they induce, rather than merely scaling model capacity or expanding data for common effects.
Recent vision-language models (VLMs)~\cite{qwen2.5,team2025kimi,bai2025qwen3,zeng2026glm} demonstrate strong visual-semantic reasoning capabilities supported by broad world knowledge, making them well suited to modeling diverse object-effect relations in real-world scenes.
To this end, we introduce \textbf{\methodname{}}, a semantic-reasoning-enhanced object-effect removal framework for high-fidelity and generalizable object removal in complex real-world videos. It is further equipped with motion-aware spatiotemporal consistency mechanisms for dynamic scenes.
To bridge high-level semantic reasoning and realistic video restoration, \methodname{} employs a \textit{VLM-based Object-Effect Reasoner} alongside a \textit{DiT-based Video Eraser} with semantic guidance.
Guided by a structured effect-analysis prompt, the \textit{Reasoner} analyzes the video input and employs learnable effect queries to aggregate removal-relevant semantic context into compact effect-aware context tokens.
Conditioned on these high-level semantic tokens together with fine-grained visual cues from the source video and mask, the \textit{Eraser} jointly removes the target and its associated effects while faithfully restoring the affected regions.
Together, this reasoning-guided generation design enables \methodname{} to better model object-induced scene changes and perform reliable removal in complex real-world scenarios.

%% P4 动态
To further improve removal stability in videos involving object motion and scene dynamics, we enhance \methodname{} with motion-aware spatiotemporal consistency mechanisms from both removal guidance and training supervision.
Specifically, motion-aware mask guidance preserves complete target coverage during temporal mask downsampling, while a motion-consistency loss encourages consistent restoration across neighboring frames, reducing missed removals, residual traces, and temporal flickerings.

%% P5数据
Training such a semantic-reasoning-enhanced framework requires data that capture diverse and complex object-scene interactions.
However, existing datasets primarily focus on a small set of predefined effects~\cite{miao2026rose}, providing limited coverage of complex real-world scenarios.
To bridge this gap, we construct the \textbf{\datasetname{}} dataset, a complex object-effect video corpus based on carefully designed 3D scenes and physics-based rendering in Unreal Engine (UE)~\cite{epicgames2026unreal}.
It complements existing data with compositional, dynamic, spatially detached, and other long-tail physical effects, such as ripples, wakes, splashes, smoke, dust, and motion trails.
Using both common and complex-effect data, we introduce a progressive training curriculum.
The model first learns basic object erasure and background restoration, and is then gradually exposed to complex effects and dynamic scenes, facilitating optimization while improving generalization and spatiotemporal robustness.

%% 总结
To summarize, our main contributions are as follows:
\begin{itemize}
    \item We propose \textbf{\methodname{}}, a semantic-reasoning-enhanced object-effect removal framework for complex real-world videos that integrates VLM-based semantic reasoning with DiT-based video restoration. Motion-aware mask guidance and a motion-consistency loss further promote spatiotemporally consistent removal in dynamic scenes.
    
    \item We construct the \textbf{\datasetname{}} dataset, a carefully designed object-effect video dataset for complex real-world removal scenarios, covering compositional, dynamic, spatially detached, and long-tail physical effects. A progressive training curriculum is further introduced to fully leverage both conventional data and \datasetname{} data, facilitating the transition from basic removal learning to complex-scene generalization.
        
    \item Extensive experiments on standard benchmarks and challenging complex-effect settings demonstrate state-of-the-art removal quality, stronger robustness to complex effects, and improved spatiotemporal stability.
\end{itemize}

\begin{figure*}[t]
    \centering
    \includegraphics[width=1.0\linewidth]{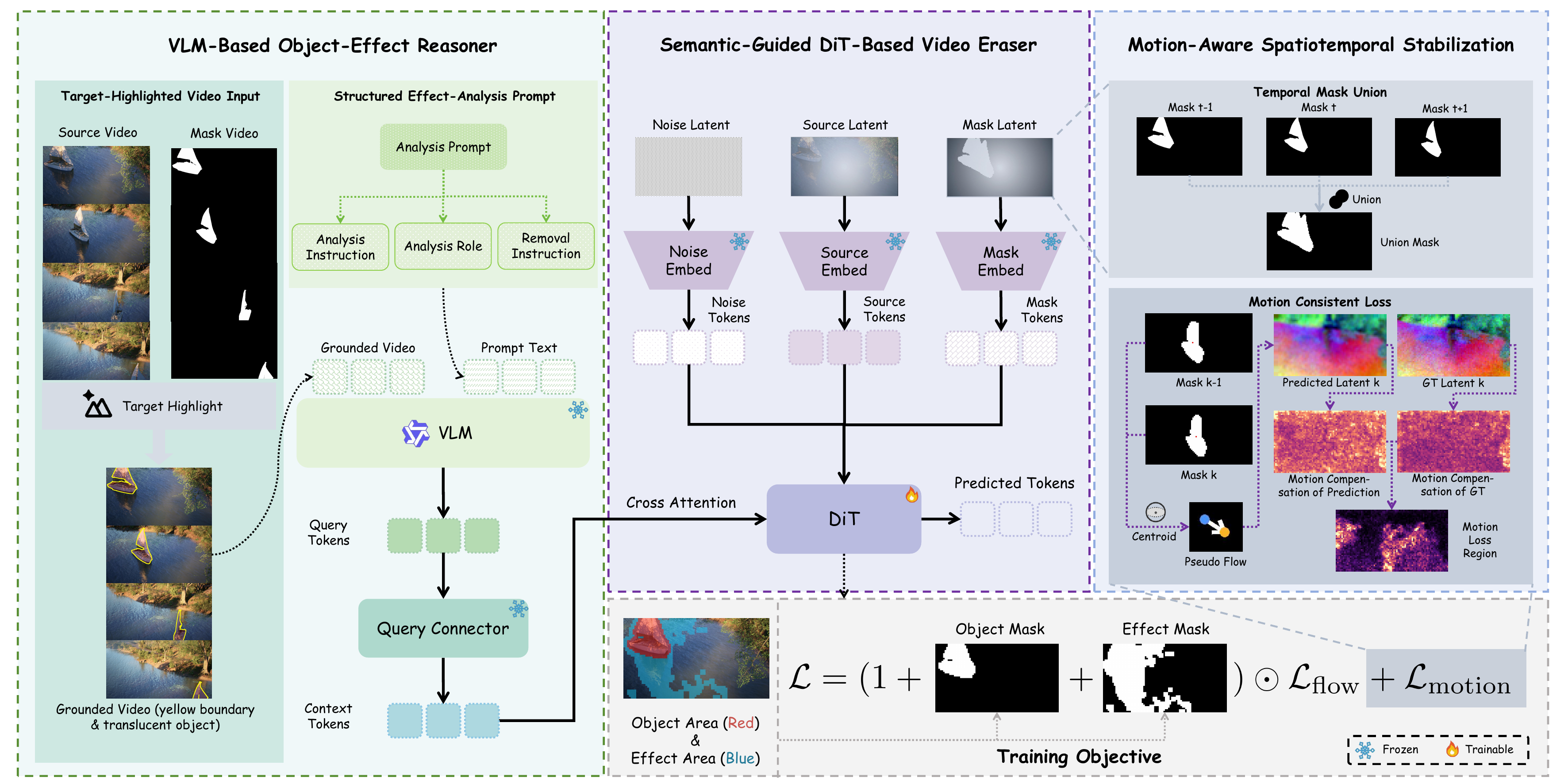}
    \caption{
    Overview of \textbf{\methodname{}}. The framework consists of two core components: 
    (1) the \textbf{VLM-Based Object-Effect Reasoner} (left), which performs cross-modal reasoning over a target-highlighted input video under structured textual guidance to extract compact semantic context; and 
    (2) the \textbf{DiT-Based Video Eraser} (middle), which integrates the semantic context with the source video and object mask to achieve high-fidelity object-effect removal. 
    Motion-aware mask guidance and a motion consistency loss (right) are further introduced to enhance spatiotemporal consistency.
    }
    % with Semantic Guidance
    \label{fig:fig1} 
\end{figure*}

\section{Related Work}
\label{related_work}
\paragraph{Video Object Removal.}
% ROSE~\cite{miao2026rose}.
% EffectErase~\cite{fu2026effecterase}.
% SVOR~\cite{hu2026ideal}.
% GenEraser~\cite{chen2026generaser}.
Video object removal aims to eliminate undesired objects while preserving visual fidelity and temporal consistency~\cite{chang2019vornet}.
Early approaches mainly rely on video inpainting techniques, where propagation-based methods~\cite{zhou2023propainter,zeng2020learning,liu2021fuseformer,li2022towards} exploit temporal correspondence for missing-region reconstruction. 
% Recent diffusion-based video generation methods further improve restoration quality by modeling complex spatiotemporal distributions~\cite{li2025diffueraser}. 
However, these approaches mainly focus on removing explicitly masked objects and often overlook object-induced effects, such as shadows, reflections, and illumination changes. 
Recent effect-aware methods attempt to address this limitation. ROSE~\cite{miao2026rose} introduces object-effect removal with dedicated supervision for side effects, while EffectErase~\cite{fu2026effecterase} explores joint object removal and insertion through reciprocal learning. 
SVOR~\cite{hu2026ideal} improves robustness under imperfect masks and real-world degradations, and GenEraser~\cite{chen2026generaser} enhances generalization through multimodal guidance and decoupled localization-preservation. 
Despite these advances, existing methods still largely rely on dataset-specific object-effect correlations and lack explicit reasoning about object-scene interactions, limiting their ability to handle complex and long-tail effects in open-world scenarios.

\paragraph{VLM Guidance in Video Editing.}
% Kiwi-Edit~\cite{lin2026kiwi}.
% Void~\cite{motamed2026void}.
% RACCOON~\cite{yoon2025raccoon}.
% Viva~\cite{cong2026viva}.
Recent VLM-based video editing methods leverage multimodal reasoning for more controllable generation. 
RACCOON~\cite{yoon2025raccoon}, Kiwi-Edit~\cite{lin2026kiwi}, Void~\cite{motamed2026void}, and Viva~\cite{cong2026viva} explore instruction-based editing, reference-guided manipulation, object interaction deletion, and reward-based alignment, respectively. 
Nevertheless, these methods mainly target explicit editing instructions or object-level manipulation, while lacking reasoning about implicit object-scene interactions and induced visual effects. 
Our \methodname{} employs the VLM as object-effect reasoners to provide semantic guidance for generalized object and effect removal.

\section{Method}
\label{method}
Given a source video $\mathbf{V}_\text{src}\in\mathbb{R}^{T\times H\times W\times 3}$ and a corresponding target-object mask $\mathbf{M} \in\{0,1\}^{T\times H\times W}$, video object removal aims to generate an edited video $\hat{\mathbf{V}}_\text{tgt}$, in which both the target object and its induced effects are removed, while the affected regions are reconstructed with high visual fidelity and spatiotemporal coherence.

Fig.~\ref{fig:fig1} illustrates our proposed \textbf{\methodname{}},
a semantic-reasoning-enhanced object-effect removal framework for complex real-world videos, further equipped with motion-aware spatiotemporal consistency mechanisms for dynamic scenes. 
% \hxnote{described as `motion-aware spatiotemporal consistency mechanisms'; `alignment' is not suitable; modify following description to match it}
%
At its core, \methodname{} couples a \textit{VLM-based Object-Effect Reasoner} and a \textit{DiT-based Video Eraser}, where the \textit{Reasoner} analyzes object-scene interactions and extracts effect-aware semantic context (Sec.~\ref{sec:method_vlm}), which guides the \textit{Eraser} to jointly remove the target and its associated effects while faithfully reconstructing the affected regions
(Sec.~\ref{sec:method_dit}).
To promote spatiotemporally consistent removal in dynamic scenes, we further introduce complementary motion-aware designs for both removal guidance and training supervision, consisting of motion-aware mask guidance and a motion-consistency loss (Sec.~\ref{sec:method_motion}).
To support the training of \methodname{} for complex real-world scenes, we develop a paired-video construction pipeline based on UE rendering and construct the \textbf{\datasetname{}} dataset, which augments existing datasets with complex object-effect cases.
A progressive training curriculum is further introduced to fully leverage both conventional and \datasetname{} data, facilitating the transition from basic removal learning to complex-scene generalization (Sec.~\ref{sec:method_data}).

\subsection{Preliminaries}
\label{sec:preliminary}

\paragraph{Video Generation Backbone.}
We adopt Wan2.2-TI2V-5B~\cite{wan2025} as our video generation backbone.
It follows the latent diffusion paradigm, where a causal 3D variational autoencoder (VAE) compresses videos into latent tokens, which are subsequently modeled by a Diffusion Transformer (DiT).
Each DiT block comprises spatiotemporal self-attention to model interactions among video tokens, cross-attention to incorporate text conditions, and a feed-forward network.

During training, given a target video latent $\mathbf{z}_0$,
we sample Gaussian noise
$\boldsymbol{\epsilon}\sim\mathcal{N}(\mathbf{0},\mathbf{I})$
and construct the noisy latent $\mathbf{z}_t$ along a linear interpolation path:
\begin{equation}
\mathbf{z}_t
=
(1-t)\mathbf{z}_0
+
t\boldsymbol{\epsilon},
\qquad
t\sim\mathcal{U}(0,1).
\label{eq:flow_path}
\end{equation}
The DiT backbone is trained with a flow-matching objective
$\mathcal{L}_{\mathrm{flow}}$ to predict the corresponding velocity field
$\boldsymbol{\epsilon}-\mathbf{z}_0$:
\begin{equation}
\mathcal{L}_{\mathrm{flow}}
=
\mathbb{E}_{t,\mathbf{z}_0,\boldsymbol{\epsilon}}
\left[
\left\|
\mathbf{v}_{\theta}(\mathbf{z}_t,t)
-
(\boldsymbol{\epsilon}-\mathbf{z}_0)
\right\|_2^2
\right].
\label{eq:flow_loss}
\end{equation}

\paragraph{VLM Backbone.}
We adopt Qwen2.5-VL-3B-Instruct~\cite{qwen2.5} as our VLM backbone. 
It integrates a native dynamic-resolution Vision Transformer with a decoder-only language model in a unified autoregressive framework. 
The vision encoder represents images and videos as spatiotemporal tokens and combines window attention with sparse global attention to efficiently capture both fine-grained details and global context. 
A visual merger compresses and projects these features into the language embedding space, where visual and textual tokens are jointly processed by the language decoder. 
% Moreover, multimodal rotary positional embeddings model temporal and two-dimensional spatial relationships, while absolute temporal encoding preserves real-time information for video understanding.
% It is an autoregressive multimodal Transformer that accepts both text and video inputs. Video frames are encoded by its vision tower and jointly modeled with the text tokens.

% \subsection{EffectLearner Framework}
% \label{sec:method_framework}

\subsection{VLM-Based Object-Effect Reasoner}
\label{sec:method_vlm}
The VLM-based Object-Effect Reasoner derives compact high-level semantic conditions for object-effect removal from multimodal inputs.
To be specific, given an input video with the removal target and a removal-analysis instruction, the VLM grounds the target and reasons about its interactions with the surrounding scene, producing hidden states that capture removal-relevant multimodal context.
To aggregate the rich VLM context into compact conditions for the video DiT, we introduce learnable effect queries
$\mathbf{Q}_{\mathrm{effect}}\in\mathbb{R}^{N_q\times d_q}$,
where $N_q$ and $d_q$ denote the number and dimension of the query tokens, respectively.
The effect queries interact with the VLM hidden states to extract removal-relevant semantics, including the target identity and motion, its induced effects and temporal evolution, the expected post-removal state, and the scene content to be preserved.
The resulting query features are then projected by a query connector from the VLM representation space to the conditioning space of the video DiT, yielding compact effect-aware context tokens $\mathbf{C}_{\mathrm{effect}}\in\mathbb{R}^{N_c\times d_c}$.
These tokens subsequently serve as high-level semantic conditions for the DiT-Based Video Eraser.

\paragraph{Target-Highlighted Video Input.}
For the visual input, we construct a target-highlighted video that helps the VLM localize the removal target while preserving its appearance and surrounding scene context.
Given a source video
$\mathbf{V}_{\mathrm{src}}\in\mathbb{R}^{T\times H\times W\times3}$
and a binary target-object mask
$\mathbf{M}\in\{0,1\}^{T\times H\times W}$,
a target-highlighted video $\mathbf{V}_{\mathrm{hl}}$ is defined as:
\begin{equation}
\mathbf{V}_{\mathrm{hl}}
=
(1-\alpha\mathbf{M})\odot\mathbf{V}_{\mathrm{src}}
+
\alpha\mathbf{M}\odot\mathbf{c}_{\mathrm{hl}},
\label{eq:target_highlight}
\end{equation}
where $\mathbf{c}_{\mathrm{hl}}\in\mathbb{R}^{3}$ denotes the highlight color,
$\alpha\in[0,1]$ is the blending coefficient,
and $\odot$ denotes element-wise multiplication, with the mask broadcast over the color channels.
This alpha-blended representation makes the target location explicit while retaining the visual evidence needed to analyze its appearance and interactions with the surrounding scene. 
% \hxnote{comparisons with separate input not described here; describe and analyze it in exp. }

\paragraph{Structured Effect-Analysis Prompt.}
For the textual input, we design a structured effect-analysis prompt that guides the VLM to analyze the highlighted target, its motion and scene interactions, the induced effects and their temporal evolution, the expected post-removal state, and the content to be preserved.
By organizing these cues into explicit analysis dimensions, the prompt encourages the VLM to examine the visual context comprehensively and identify subtle or weakly correlated effects in a systematic manner.
The prompt is formulated as:
\begin{equation}
P=[P_{\mathrm{sys}};P_{\mathrm{inst}};P_{\mathrm{task}}],
\label{eq:analysis_prompt}
\end{equation}
where $P_{\mathrm{sys}}$, $P_{\mathrm{inst}}$, and $P_{\mathrm{task}}$
denote the analysis role, the general effect-analysis instruction, and the sample-level removal instruction, respectively.
The complete prompt is provided in \suppl.

\begin{figure}[t]
    \centering
    \includegraphics[width=1.0\linewidth]{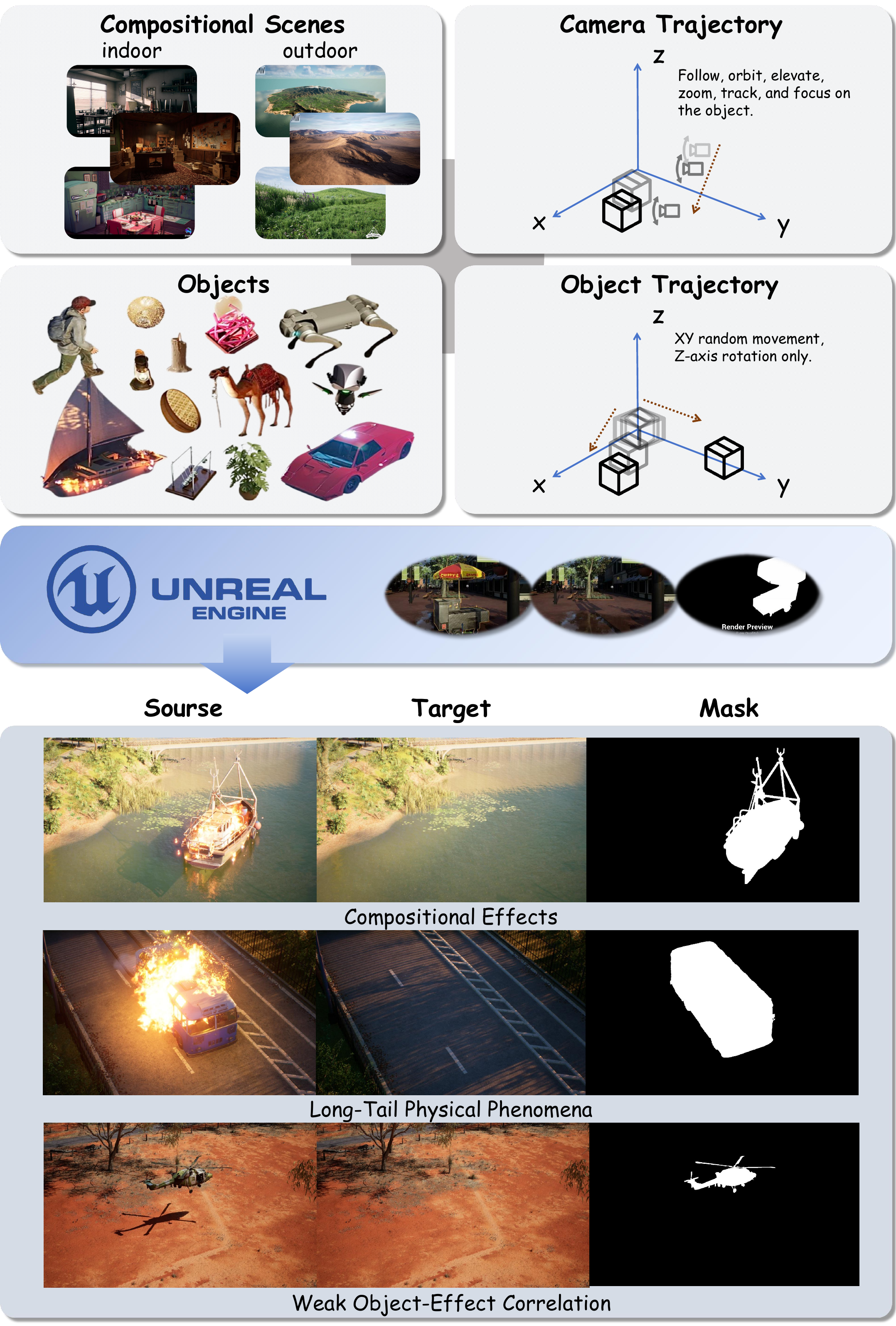}
    \caption{Data construction pipeline of \textbf{\datasetname{}}. }
    \label{fig:fig2} 
\end{figure}

\subsection{Semantic-Guided DiT-Based Video Eraser}
\label{sec:method_dit}
Given the effect-aware semantic context extracted by the VLM, we employ a video DiT to perform complete and high-quality object removal using both low-level visual conditions and high-level semantic guidance.
Specifically, the source video $\mathbf{V}_{\mathrm{src}}$ is encoded by the VAE into the source latent $\mathbf{z}_{\mathrm{src}}$, while the object mask $\mathbf{M}_{\mathrm{obj}}$ is downsampled to $\bar{\mathbf{M}}_{\mathrm{obj}}$ to match the spatiotemporal resolution of the video latent.
The source latent $\mathbf{z}_{\mathrm{src}}$ and the downsampled mask $\bar{\mathbf{M}}_{\mathrm{obj}}$ are concatenated with the noisy target latent $\mathbf{z}_t$ along the channel dimension and fed into the DiT as low-level visual conditions.
The source latent preserves the appearance, texture, structure, and temporal information of the input video, while the mask specifies the target region for removal.
Meanwhile, the effect-aware context tokens $\mathbf{C}_{\mathrm{effect}}$ replace the original text condition in the DiT cross-attention layers, providing high-level semantic guidance about the target, its induced effects, the expected post-removal state, and the scene content to be preserved.

To emphasize removal-critical regions, particularly the target object and its induced effects, we introduce a region-weighted flow-matching objective.
The joint object-effect mask is obtained by thresholding the absolute difference between the paired source and target videos:
\begin{equation}
\mathbf{M}_{\mathrm{effect}}
=
\mathbf{1}
\left[
\left|
\mathbf{V}_{\mathrm{src}}
-
\mathbf{V}_{\mathrm{tgt}}
\right|
>
\tau
\right],
\label{eq:effect_mask}
\end{equation}
where $\tau$ denotes the difference threshold and $\mathbf{1}[\cdot]$ is the indicator function.
Using the object mask $\mathbf{M}_{\mathrm{obj}}$ and the effect mask $\mathbf{M}_{\mathrm{effect}}$, the weighted flow-matching objective is defined as:
\begin{equation}
\mathcal{L}_{\mathrm{w\text{-}flow}}
=
\left(
1
+
w_{\mathrm{obj}}\mathbf{M}_{\mathrm{obj}}
+
w_{\mathrm{effect}}\mathbf{M}_{\mathrm{effect}}
\right)
\odot
\mathcal{L}_{\mathrm{flow}},
\label{eq:weighted_flow_loss}
\end{equation}
where $w_{\mathrm{obj}}$ and $w_{\mathrm{effect}}$ control the additional weights assigned to the object and effect regions, respectively.
This objective encourages the model to focus on both the explicitly specified object region and the broader regions affected by its induced effects.

\subsection{Motion-Aware Spatiotemporal Stabilization}
\label{sec:method_motion}

Beyond semantic reasoning, robust video object removal also requires stable restoration in dynamic scenes.
For example, moving targets may occupy substantially different locations across neighboring frames, while their induced effects may evolve over time.
However, temporal mask compression can discard short-lived target regions, and insufficient cross-frame supervision may further lead to inconsistent restoration, resulting in incomplete removal, residual traces, and flicker.
To address these issues, we introduce motion-aware spatiotemporal stabilization through complementary designs for removal guidance and training supervision.

% \paragraph{Motion-Aware Mask Guidance.}
% Due to temporal compression in the VAE, multiple video frames are mapped into a single latent temporal position. 
% Directly using frame-level masks may fail to cover the complete trajectory of moving objects, especially under fast motion.
% Therefore, we construct a \textit{motion-aware temporal union mask} to preserve the complete removal region within each latent temporal window.
% Specifically, let $M=\{M^{(i)}\}_{i=1}^{T}$ denote the input binary object-mask sequence, where $M^{(i)}$ indicates the target region in the $i$-th frame.
% For the $k$-th latent temporal position, we collect the corresponding input frames $\Omega_k$ and compute:
% \begin{equation}
% M_{\mathrm{union}}^k
% =
% \max_{i\in\Omega_k}M^{(i)},
% \label{eq:temporal_union}
% \end{equation}
% where the $\max$ operation is performed element-wise.
% The obtained union mask provides more complete spatial guidance for object removal.
% We further dilate the mask to obtain the supervision region.
% % :
% % \begin{equation}
% % M_{\mathrm{loss}}^k
% % =
% % \mathrm{Dilate}(M_{\mathrm{union}}^k,r_m),
% % \label{eq:dilate_mask}
% % \end{equation}
% % where $r_m$ denotes the dilation radius, controlling the expansion range around the target boundary. 
% This operation compensates for possible spatial misalignment caused by temporal compression and mask downsampling, while covering surrounding regions affected by object removal.
% $M_{\mathrm{union}}$ is used as the DiT conditioning mask, while $M_{\mathrm{loss}}$ is used to construct the region-aware training objective. 

\begin{table*}[t]
\centering
\small
\setlength{\tabcolsep}{3pt}
\renewcommand{\arraystretch}{0.9}
\resizebox{\textwidth}{!}{
\begin{tabular}{lccccc|ccccc}
\toprule
\multirow{2}{*}{Method}
& \multicolumn{5}{c|}{ROSE-Bench}
& \multicolumn{5}{c}{\ourbencheval{}} \\
& PSNR$\uparrow$ & SSIM$\uparrow$ & LPIPS$\downarrow$ & MAE$\downarrow$ & FVD$\downarrow$
& PSNR$\uparrow$ & SSIM$\uparrow$ & LPIPS$\downarrow$ & MAE$\downarrow$ & FVD$\downarrow$ \\
\midrule
ProPainter~\cite{zhou2023propainter}
& 26.661 & 0.917 & 0.096 & 8.208 & 174.795
& 25.437 & \underline{0.897} & 0.127 & \underline{9.628} & 364.697 \\
DiffuEraser~\cite{li2025diffueraser}
& 25.384 & 0.885 & 0.112 & 10.214 & 168.563
& 24.587 & 0.867 & 0.153 & 11.207 & 342.111 \\
VACE~\cite{jiang2025vace}
& 21.345 & 0.782 & 0.158 & 13.641 & 293.012
& 17.745 & 0.544 & 0.301 & 23.534 & 783.915 \\
ROSE~\cite{miao2026rose}
& \textbf{30.468} & \textbf{0.933} & \underline{0.058} & \textbf{6.002} & 803.722
& \underline{25.955} & 0.882 & \underline{0.111} & 10.476 & 887.541 \\
EffectErase~\cite{fu2026effecterase}
& 27.012 & 0.916 & 0.085 & 8.807 & \underline{135.117}
& 24.139 & 0.857 & 0.128 & 13.580 & \underline{238.521} \\
\methodname{} (Ours)
& \underline{28.973} & \underline{0.921} & \textbf{0.054} & \underline{6.781} & \textbf{81.866}
& \textbf{29.521} & \textbf{0.934} & \textbf{0.044} & \textbf{6.839} & \textbf{69.184} \\
\bottomrule
\end{tabular}}
\caption{Quantitative results on ROSE-Bench and \ourbencheval{}. The best and second-best results are \textbf{bolded} and \underline{underlined}.}
\label{tab:1}
\end{table*}

\paragraph{Motion-Aware Mask Guidance.}
Due to temporal compression in the VAE, multiple video frames are mapped into a single latent temporal position, making frame-level masks insufficient to capture the complete motion trajectory under fast object movement.
Inspired by MUSE in~\cite{hu2026ideal}, we construct a \textit{motion-aware temporal union mask} to provide more complete spatial guidance within each latent temporal window.
Specifically, given the binary object-mask sequence $M=\{M^{(i)}\}_{i=1}^{T}$, we aggregate the masks corresponding to the $k$-th latent temporal position $\Omega_k$ as:
\begin{equation}
M_{\mathrm{union}}^k
=
\max_{i\in\Omega_k}M^{(i)},
\label{eq:temporal_union}
\end{equation}
where the maximum operation is performed element-wise.
The resulting union mask covers the full removal trajectory and is further dilated to obtain the supervision region, compensating for spatial misalignment caused by temporal compression and mask downsampling.

\paragraph{Motion Consistency Supervision.}
While the region-weighted flow-matching objective $\mathcal{L}_{\mathrm{w\text{-}flow}}$ improves spatial restoration, it does not explicitly enforce temporal consistency across adjacent frames.
To address temporal artifacts caused by dynamic object movement and effect evolution, we introduce a \textit{motion consistent loss}.
Specifically, we estimate the motion between adjacent latent frames based on the displacement of the removal region and construct a translational pseudo-flow $f_k$ from frame $k-1$ to frame $k$.
Given the backward warping operator $\mathcal{W}(\cdot,f_k)$, we align the previous latent representation to the current frame and enforce temporal consistency between the predicted and target latent dynamics:
\begin{equation}
\begin{aligned}
\hat{\Delta}_k = \hat{x}_0^k-\mathcal{W}(\hat{x}_0^{k-1},f_k), \qquad
\Delta_k = x_0^k-\mathcal{W}(x_0^{k-1},f_k),\\
\mathcal{L}_{\mathrm{motion}} = \operatorname{Mean}_{R_k}
\left\|
\hat{\Delta}_k-\Delta_k
\right\|_2^2,\qquad\qquad
\end{aligned}
\label{eq:motion_loss}
\end{equation}
where $\hat{x}_0^k$ and $x_0^k$ denote the predicted and ground-truth clean latent representations at the $k$-th frame, respectively;
$\hat{\Delta}_k$ and ${\Delta}_k$ denote motion-compensated temporal change of the prediction and the ground-truth target, respectively; 
$R_k$ represents the temporal supervision region derived from the removal-related regions.
By matching motion-compensated latent changes, $\mathcal{L}_{\mathrm{motion}}$ encourages temporally consistent restoration of the removed object and its induced effects.
% The warping operator $\mathcal{W}(\cdot,f_k)$ spatially aligns the previous-frame latent with the current frame according to the estimated motion. Accordingly, $\hat{\Delta}_k$ and $\Delta_k$ represent the motion-compensated temporal changes in the predicted and ground-truth latents, respectively. By matching these residual changes, $\mathcal{L}_{\mathrm{motion}}$ encourages the restored content to follow the target temporal evolution rather than simply enforcing identical adjacent frames. Finally, $R_k$ denotes the supervision region derived from the removal regions in the adjacent frames, restricting the loss to areas associated with the removed object and its induced effects.
% where $R_k$ denotes the supervision region derived from the removal region.
% The motion consistency loss encourages the restored object-free regions and induced effects to evolve consistently across frames.
The final training objective is:
\begin{equation}
\mathcal{L}
=
\mathcal{L}_{\mathrm{w-flow}}
+
\lambda_{\mathrm{motion}}
\mathcal{L}_{\mathrm{motion}},
\label{eq:final_loss}
\end{equation}
where $\lambda_{\mathrm{motion}}$ controls the temporal consistency constraint.

\subsection{\datasetname{} Dataset and Progressive Training Curriculum}
\label{sec:method_data}
%%%%%%%%%%%%% data pipeline 新描述
To complement existing VOR data primarily covering common object-removal scenarios with simple effects, we construct the \datasetname{} dataset comprising complex object-effect videos to provide targeted supervision for challenging object-induced effects, as illustrated in Figure~\ref{fig:fig2}. 
We first compose diverse indoor and outdoor environments with a broad collection of foreground objects, covering humans, animals, vehicles, plants, and various object categories. 
To increase spatiotemporal diversity, we independently control camera and object motion. 
The camera follows target-centric trajectories that combine following, orbiting, elevation, zooming, tracking, and refocusing operations, while the target undergoes random translation on the horizontal plane and rotation around the vertical axis. 
Given each configured scene, UE renders three spatially and temporally aligned sequences under identical scene states and trajectories: a source video containing the target and its object-induced effects, a target video in which the target and its causally generated effects are removed, and a binary mask video identifying only the target region. 
This controllable rendering process enables accurate supervision for challenging cases involving compositional effects, uncommon physical effects, and weakly correlated or spatially detached object-effect relationships.

Also, we organize training as a difficulty-progressive curriculum. The model first learns basic object removal, common induced effects, and background restoration from standard paired supervision. We then increase the proportion of compositional, dynamic, and spatially detached samples, followed by refinement on long-tail physical effects and fast-motion cases. This progression moves the training objective from basic removal toward complex effect reasoning and temporal stabilization.

\begin{table*}[t]
\centering
% \small
\normalsize
\setlength{\tabcolsep}{6pt}
\begin{tabular}{lcccccc}
\toprule
Method
& \makecell{Subject\\Consistency}
& \makecell{Background\\Consistency}
& \makecell{Motion\\Smoothness}
& \makecell{Dynamic\\Degree}
& \makecell{Imaging\\Quality}
& \makecell{Total\\Score}\\
\midrule
ProPainter~\cite{zhou2023propainter}
& 0.948 & 0.955 & \textbf{0.986} & \underline{0.386} & 0.534 & 0.762 \\
DiffuEraser~\cite{li2025diffueraser}
& 0.955 & \underline{0.957} & \textbf{0.986} & \underline{0.386} & 0.545 & \underline{0.766} \\
VACE~\cite{jiang2025vace}
& 0.949 & 0.945 & 0.984 & \underline{0.386} & \textbf{0.593} & \textbf{0.772}  \\
ROSE~\cite{miao2026rose}
& 0.945 & 0.949 & \underline{0.985} & 0.371 & 0.565 & 0.763 \\
EffectErase~\cite{fu2026effecterase}
& \textbf{0.965} & 0.953 & \textbf{0.986}& 0.343 & \underline{0.575} & 0.765 \\
\methodname{} (Ours)
& \underline{0.958} & \textbf{0.958} & \textbf{0.986} & \textbf{0.400} & 0.557 & \textbf{0.772} \\
\bottomrule
\end{tabular}
\caption{Quantitative results of various methods in \ourbenchwild{}. The best and second-best results are \textbf{bolded} and \underline{underlined}.}
\label{tab:2}
\end{table*}

\section{Experiments}
\label{exp}
% We evaluate \methodname{} from three complementary perspectives. ROSE-Bench measures performance on an established paired benchmark. EffectWorld diagnoses object--effect removal under compositional, spatially detached, long-tail, and dynamic conditions. Finally, \ourbenchwild evaluates real-world generalization without relying on the UE rendering domain. All reproduced methods receive the same source video and binary object mask, and their outputs are aligned to the same temporal range, spatial resolution, and frame rate before evaluation.

\subsection{Experimental Settings}
% \TODO{more details.}
\boldparagraph{Training Data.}
We employ 16,663 ROSE triplets and 11,048 \datasetname{} triplets during training, which serve as common and complex scenarios, respectively.

\boldparagraph{Benchmarks.}
We evaluate models on ROSE-Bench~\cite{miao2026rose} as well as our \ourbencheval{} and \ourbenchwild{}. 
% \TODO{Some details of each bench.}
ROSE-Bench contains 60 synthetic video triplets.
\ourbencheval{} contains 33 paired complex-effect cases.
\ourbenchwild{} contains 70 curated open-world videos, including 59 manually reviewed videos from the Pexels community~\cite{pexels2026videos} and 11 videos from DAVIS~\cite{perazzi2016benchmark}.
Details can be found in \suppl.

\boldparagraph{Metrics.}
For ROSE-Bench and \ourbencheval{}, we report PSNR, SSIM, LPIPS, MAE, and FVD to align with previous video inpainting methods. 
% \TODO{Vbench Metrics introduction for \ourbenchwild.}
For \ourbenchwild, we apply VBench~\cite{huang2024vbench} to evaluate models from Subject Consistency, Background Consistency, Motion Smoothness, Dynamic Degree, and Imaging Quality.
% Details can be found in \suppl.

%%%%%%%%%%%%5 实验配置放附录！
\boldparagraph{Implementation Details.}
% \TODO{hyperparameters should be checked again.}
For the VLM backbone, we employ Qwen2.5-VL-3B-Instruct~\cite{qwen2.5}. 
% 下面这个放附录？？？感觉写不下
% Each $1280\times704$ target-highlighted video is processed with aspect-ratio-preserving resizing and model-aligned rounding to $672\times364$ under a maximum per-frame pixel budget of 262,144.
% By default, we sample videos at 2 fps and uniformly subsample them to a maximum of 16 frames if the resulting frame count exceeds 16.
For the DiT backbone, we employ Wan2.2-TI2V-5B~\cite{wan2025}, initialized from Kiwi-Edit~\cite{lin2026kiwi}, and fine-tune only the DiT blocks on 89-frame clips at $1280\times704$ resolution. We train for 1 epoch on NVIDIA L20 GPUs with a global batch size of 12, using AdamW with a learning rate of $5\times10^{-6}$. During inference, we use 50 denoising steps.

\begin{figure*}[t]
    \centering
    \includegraphics[width=1.0\linewidth]{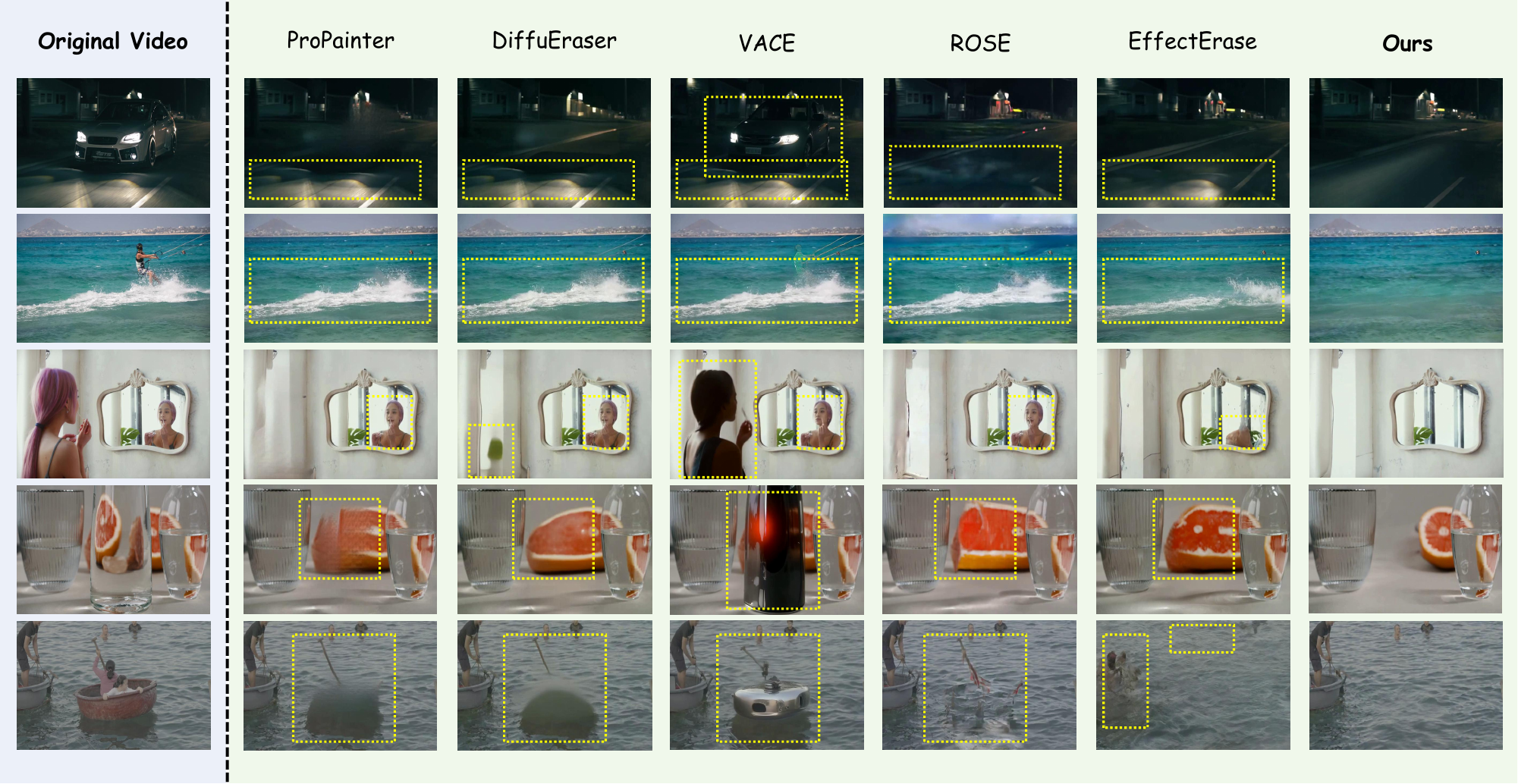}
    \caption{Visualization results of various methods on \textbf{\ourbenchwild}. Yellow dashed boxes denote regions with incompletely removed effects, incompletely erased target objects, or inconsistent background content.}
    % with Semantic Guidance
    \label{fig:fig3} 
\end{figure*}

\subsection{Main Results}
\label{sec:main_results}

\noindent\textbf{Results on ROSE-Bench and EffectWorld-Eval.}
Table~\ref{tab:1} highlights the differences between frame-level fidelity, temporal coherence, and cross-domain generalization. On ROSE-Bench, ROSE performs strongly on pixel-level metrics due to its category-specific supervision, but its high FVD indicates limited video-level coherence. In contrast, \methodname{} achieves competitive reconstruction quality together with the best LPIPS and substantially lower FVD, demonstrating a better perceptual and temporal balance.
Traditional inpainting remains competitive, while VACE shows that general editing ability does not ensure precise object-effect removal.
On the more challenging EffectWorld-Eval, competing methods degrade considerably because conventional inpainting lacks effect reasoning, while existing effect-aware models remain tied to predefined categories. \methodname{} consistently performs strongly across all metrics, particularly LPIPS and FVD, demonstrating improved generalization to compositional, dynamic, spatially detached, and long-tail effects.

\noindent\textbf{Results on EffectWorld-Wild.}
Table~\ref{tab:2} reveals distinct behaviors among the compared methods. Traditional inpainting approaches, such as ProPainter and DiffuEraser, preserve stable subjects and backgrounds but exhibit limited dynamics, suggesting conservative restoration. VACE achieves stronger frame-level Imaging Quality at the cost of lower Background Consistency. Among effect-aware methods, EffectErase obtains the highest Subject Consistency but the lowest Dynamic Degree, indicating possible temporal over-smoothing, while ROSE shows limited generalization from predefined effects to open-world interactions.
In contrast, \methodname{} achieves a more balanced trade-off, combining the strongest Background Consistency and Dynamic Degree with competitive Subject Consistency and Motion Smoothness. This indicates that it preserves unrelated content while retaining natural dynamics. 
In general, \methodname{} achieves the best total score through stronger background and temporal behavior, demonstrating the effectiveness of semantic object-effect reasoning and motion-aware stabilization.

\subsection{Visualization Results}
Figure~\ref{fig:fig3} presents qualitative comparisons on challenging real-world videos from \ourbenchwild{}. Traditional inpainting methods, such as ProPainter and DiffuEraser, mainly respond to the explicitly masked target region and therefore often leave spatially extended or detached effects in the scene; their reconstructed regions may also exhibit blurring or visible appearance discontinuities. General-purpose video editing models such as VACE provide stronger generative completion, but tend to alter unrelated scene content or hallucinate structures when precise restoration is required. Although ROSE and EffectErase better handle common effects, they still produce residual traces or excessive removal in cases involving complex, dynamic, and weakly correlated object--effect d. In contrast, \methodname{} more consistently removes both the target and its complete causal influence while preserving surrounding objects, background structures, and natural scene dynamics. The restored regions exhibit fewer artifacts and better integration with adjacent content, demonstrating that semantic object--effect reasoning improves removal completeness, while motion-aware stabilization promotes coherent restoration in open-world videos.

\begin{table}[t]
\centering
\small
\setlength{\tabcolsep}{3pt}

\begin{tabular}{l|ccccc}
\toprule
\multirow{2}{*}{Variant}
& \multicolumn{5}{c}{\ourbencheval{}} \\
& PSNR$\uparrow$ & SSIM$\uparrow$ & LPIPS$\downarrow$ & MAE$\downarrow$ & FVD$\downarrow$ \\
\midrule
w/o VLM
& 27.737 & 0.915 & 0.069 & 8.330 & 118.236 \\
w/o $\mathcal{L}_{\mathrm{w-flow}}$
& 27.381 & 0.900 & 0.081 & 7.683 & 129.289 \\
w/o $M_{\mathrm{union}}$
& 27.648 & 0.904 & 0.077 & 7.521 & 112.326 \\
w/o $\mathcal{L}_{\mathrm{motion}}$
& 24.703 & 0.764 & 0.187 & 21.523 & 294.778 \\
Full method
& 29.521 & 0.934 & 0.044 & 6.839 & 69.184 \\
\bottomrule
\end{tabular}
\caption{Ablation results of our proposed components.}
\label{tab:method_ablation}
\end{table}

\subsection{Ablation Studies}
\label{sec:ablations}
Table~\ref{tab:method_ablation} shows that the proposed components address complementary failure modes. Removing VLM guidance degrades reconstruction and video-level metrics, indicating that object masks alone cannot identify weakly correlated or spatially detached effects. Effect-aware semantic context therefore helps the DiT determine both which target-induced changes should be removed and which content should be preserved.
Without region-weighted flow matching, background regions dominate optimization, leaving the smaller object and effect regions insufficiently supervised. Removing the temporal union mask causes a distinct degradation: temporal VAE compression may discard short-lived positions of fast-moving objects, resulting in incomplete removal or residual traces. The largest deterioration occurs without the motion-consistent loss, particularly in LPIPS and FVD, showing that spatial supervision alone cannot ensure coherent temporal evolution. Matching motion-compensated temporal residuals is thus crucial for suppressing flicker and maintaining cross-frame consistency. Overall, semantic reasoning improves effect localization, region weighting strengthens removal supervision, and the motion-aware components stabilize dynamic video restoration.

\section{Conclusion}
We presented \textbf{\methodname{}}, a semantic-reasoning-enhanced framework for removing target objects together with their induced effects from complex videos. Its VLM-based Object-Effect Reasoner converts target-highlighted visual evidence and a structured effect-analysis prompt into compact effect-aware context, which guides a DiT-based Video Eraser in identifying the scene changes that should be removed or preserved. Motion-aware mask guidance and motion-consistency supervision further improve removal coverage and temporal stability under object motion, temporally evolving effects, and fast motion. We also constructed the \textbf{\datasetname{}} dataset to complement conventional supervision with compositional effects, weakly correlated effects, and long-tail physical phenomena, and integrated these data through a progressive training curriculum. Experiments on ROSE-Bench, EffectWorld-Eval, and EffectWorld-Wild demonstrate strong reconstruction quality, perceptual fidelity, temporal coherence, and open-world generalization, supporting semantic reasoning and motion-aware stabilization as complementary components for reliable video object-and-effect removal.

% \clearpage
\bibliography{aaai2027}

@article{miao2026rose,
  title={Rose: Remove objects with side effects in videos},
  author={Miao, Chenxuan and Feng, Yutong and Zeng, Jianshu and Gao, Zixiang and Liu, Hantang and Yan, Yunfeng and Qi, Donglian and Chen, Xi and Wang, Bin and Zhao, Hengshuang},
  journal={Advances in Neural Information Processing Systems},
  volume={38},
  pages={149140--149162},
  year={2026}
}

@inproceedings{mao2026omni,
  title={Omni-effects: Unified and spatially-controllable visual effects generation},
  author={Mao, Fangyuan and Hao, Aiming and Chen, Jintao and Liu, Dongxia and Feng, Xiaokun and Zhu, Jiashu and Wu, Meiqi and Chen, Chubin and Wu, Jiahong and Chu, Xiangxiang},
  booktitle={Proceedings of the AAAI Conference on Artificial Intelligence},
  volume={40},
  number={10},
  pages={7927--7935},
  year={2026}
}

@article{liu2026mozoo,
  title={MoZoo: Unleashing Video Diffusion power in animal fur and muscle simulation},
  author={Liu, Dongxia and Ma, Jie and Yang, Xiaochen and Zhang, Jiancheng and Xia, Bin and Kan, Zhehan and Huang, Nisha and Liang, Jun and Yang, Wenming and Li, Jin},
  journal={arXiv preprint arXiv:2605.13857},
  year={2026}
}

@inproceedings{li2025realcam,
  title={Realcam-i2v: Real-world image-to-video generation with interactive complex camera control},
  author={Li, Teng and Zheng, Guangcong and Jiang, Rui and Zhan, Shuigen and Wu, Tao and Lu, Yehao and Lin, Yining and Deng, Chuanyun and Xiong, Yepan and Chen, Min and others},
  booktitle={Proceedings of the IEEE/CVF International Conference on Computer Vision},
  pages={28785--28796},
  year={2025}
}

@inproceedings{huang2025dreamphysics,
  title={Dreamphysics: Learning physics-based 3d dynamics with video diffusion priors},
  author={Huang, Tianyu and Zhang, Haoze and Zeng, Yihan and Zhang, Zhilu and Li, Hui and Zuo, Wangmeng and Lau, Rynson WH},
  booktitle={Proceedings of the AAAI Conference on Artificial Intelligence},
  volume={39},
  number={4},
  pages={3733--3741},
  year={2025}
}

@inproceedings{zhang2024physdreamer,
  title={Physdreamer: Physics-based interaction with 3d objects via video generation},
  author={Zhang, Tianyuan and Yu, Hong-Xing and Wu, Rundi and Feng, Brandon Y and Zheng, Changxi and Snavely, Noah and Wu, Jiajun and Freeman, William T},
  booktitle={European Conference on Computer Vision},
  pages={388--406},
  year={2024},
  organization={Springer}
}

@misc{openai2026gpt54,
  author       = {{OpenAI}},
  title        = {Introducing {GPT-5.4}},
  year         = {2026},
  howpublished = {\url{https://openai.com/index/introducing-gpt-5-4/}}
}

@article{zeng2026glm,
  title={Glm-5: from vibe coding to agentic engineering},
  author={Zeng, Aohan and Lv, Xin and Hou, Zhenyu and Du, Zhengxiao and Zheng, Qinkai and Chen, Bin and Yin, Da and Ge, Chendi and Huang, Chenghua and Xie, Chengxing and others},
  journal={arXiv preprint arXiv:2602.15763},
  year={2026}
}

@article{bai2025qwen3,
  title={Qwen3-vl technical report},
  author={Bai, Shuai and Cai, Yuxuan and Chen, Ruizhe and Chen, Keqin and Chen, Xionghui and Cheng, Zesen and Deng, Lianghao and Ding, Wei and Gao, Chang and Ge, Chunjiang and others},
  journal={arXiv preprint arXiv:2511.21631},
  year={2025}
}

@article{team2025kimi,
  title={Kimi-vl technical report},
  author={Team, Kimi and Du, Angang and Yin, Bohong and Xing, Bowei and Qu, Bowen and Wang, Bowen and Chen, Cheng and Zhang, Chenlin and Du, Chenzhuang and Wei, Chu and others},
  journal={arXiv preprint arXiv:2504.07491},
  year={2025}
}

@article{yu2024barriers,
  title={Barriers to industry adoption of AI video generation tools: A study based on the perspectives of video production professionals in China},
  author={Yu, Tao and Yang, Wei and Xu, Junping and Pan, Younghwan},
  journal={Applied Sciences},
  volume={14},
  number={13},
  pages={5770},
  year={2024},
  publisher={MDPI}
}

@inproceedings{perazzi2016benchmark,
  title={A benchmark dataset and evaluation methodology for video object segmentation},
  author={Perazzi, Federico and Pont-Tuset, Jordi and McWilliams, Brian and Van Gool, Luc and Gross, Markus and Sorkine-Hornung, Alexander},
  booktitle={Proceedings of the IEEE conference on computer vision and pattern recognition},
  pages={724--732},
  year={2016}
}

@misc{epicgames2026unreal,
  author       = {{Epic Games}},
  title        = {{Unreal Engine 5}},
  year         = {2020},
  howpublished = {\url{https://www.unrealengine.com/}}
}

@misc{pexels2026videos,
  title = {Free Stock Videos Shared by the Pexels Community},
  author = {Pexels},
  year = {2014},
  howpublished = {\url{https://www.pexels.com/videos/}}
}

@inproceedings{liu2021fuseformer,
  title={Fuseformer: Fusing fine-grained information in transformers for video inpainting},
  author={Liu, Rui and Deng, Hanming and Huang, Yangyi and Shi, Xiaoyu and Lu, Lewei and Sun, Wenxiu and Wang, Xiaogang and Dai, Jifeng and Li, Hongsheng},
  booktitle={Proceedings of the IEEE/CVF international conference on computer vision},
  pages={14040--14049},
  year={2021}
}

@inproceedings{zeng2020learning,
  title={Learning joint spatial-temporal transformations for video inpainting},
  author={Zeng, Yanhong and Fu, Jianlong and Chao, Hongyang},
  booktitle={European conference on computer vision},
  pages={528--543},
  year={2020},
  organization={Springer}
}

@inproceedings{chang2019vornet,
  title={Vornet: Spatio-temporally consistent video inpainting for object removal},
  author={Chang, Ya-Liang and Yu Liu, Zhe and Hsu, Winston},
  booktitle={Proceedings of the IEEE/CVF conference on computer vision and pattern recognition workshops},
  pages={0--0},
  year={2019}
}

@article{lin2026kiwi,
  title={Kiwi-edit: Versatile video editing via instruction and reference guidance},
  author={Lin, Yiqi and Liang, Guoqiang and Zeng, Ziyun and Bai, Zechen and Chen, Yanzhe and Shou, Mike Zheng},
  journal={arXiv preprint arXiv:2603.02175},
  year={2026}
}

@article{chen2026generaser,
  title={GenEraser: Generalizable Video Object Removal via Balanced Text-Mask Guidance and Decoupled Locator-Preserver},
  author={Chen, Yuqing and Liu, Lin and Wu, Haisu and Zhang, Xiaopeng and Wang, Yaowei and Yang, Yujiu and Tian, Qi},
  journal={arXiv preprint arXiv:2605.30045},
  year={2026}
}

@inproceedings{huang2024vbench,
  title={Vbench: Comprehensive benchmark suite for video generative models},
  author={Huang, Ziqi and He, Yinan and Yu, Jiashuo and Zhang, Fan and Si, Chenyang and Jiang, Yuming and Zhang, Yuanhan and Wu, Tianxing and Jin, Qingyang and Chanpaisit, Nattapol and others},
  booktitle={Proceedings of the IEEE/CVF Conference on Computer Vision and Pattern Recognition},
  pages={21807--21818},
  year={2024}
}

@inproceedings{zhou2023propainter,
  title={Propainter: Improving propagation and transformer for video inpainting},
  author={Zhou, Shangchen and Li, Chongyi and Chan, Kelvin CK and Loy, Chen Change},
  booktitle={Proceedings of the IEEE/CVF international conference on computer vision},
  pages={10477--10486},
  year={2023}
}

@inproceedings{cong2026viva,
  title={Viva: Vlm-guided instruction-based video editing with reward optimization},
  author={Cong, Xiaoyan and Yang, Haotian and Wang, Angtian and Wang, Yizhi and Yang, Yiding and Zhang, Canyu and Ma, Chongyang},
  booktitle={Proceedings of the IEEE/CVF Conference on Computer Vision and Pattern Recognition},
  pages={34364--34374},
  year={2026}
}

@inproceedings{yoon2025raccoon,
  title={RACCOON: Versatile Instructional Video Editing with Auto-Generated Narratives},
  author={Yoon, Jaehong and Yu, Shoubin and Bansal, Mohit},
  booktitle={Proceedings of the 2025 Conference on Empirical Methods in Natural Language Processing},
  pages={27960--27996},
  year={2025}
}

@misc{qwen2.5,
    title = {Qwen2.5: A Party of Foundation Models},
    url = {https://qwenlm.github.io/blog/qwen2.5/},
    author = {Qwen Team},
    month = {September},
    year = {2024}
}

@article{wan2025,
      title={Wan: Open and Advanced Large-Scale Video Generative Models}, 
      author={Team Wan and Ang Wang and Baole Ai and Bin Wen and Chaojie Mao and Chen-Wei Xie and Di Chen and Feiwu Yu and Haiming Zhao and Jianxiao Yang and Jianyuan Zeng and Jiayu Wang and Jingfeng Zhang and Jingren Zhou and Jinkai Wang and Jixuan Chen and Kai Zhu and Kang Zhao and Keyu Yan and Lianghua Huang and Mengyang Feng and Ningyi Zhang and Pandeng Li and Pingyu Wu and Ruihang Chu and Ruili Feng and Shiwei Zhang and Siyang Sun and Tao Fang and Tianxing Wang and Tianyi Gui and Tingyu Weng and Tong Shen and Wei Lin and Wei Wang and Wei Wang and Wenmeng Zhou and Wente Wang and Wenting Shen and Wenyuan Yu and Xianzhong Shi and Xiaoming Huang and Xin Xu and Yan Kou and Yangyu Lv and Yifei Li and Yijing Liu and Yiming Wang and Yingya Zhang and Yitong Huang and Yong Li and You Wu and Yu Liu and Yulin Pan and Yun Zheng and Yuntao Hong and Yupeng Shi and Yutong Feng and Zeyinzi Jiang and Zhen Han and Zhi-Fan Wu and Ziyu Liu},
      journal = {arXiv preprint arXiv:2503.20314},
      year={2025}
}

@inproceedings{jiang2025vace,
  title={Vace: All-in-one video creation and editing},
  author={Jiang, Zeyinzi and Han, Zhen and Mao, Chaojie and Zhang, Jingfeng and Pan, Yulin and Liu, Yu},
  booktitle={Proceedings of the IEEE/CVF International Conference on Computer Vision},
  pages={17191--17202},
  year={2025}
}

@article{li2025diffueraser,
  title={Diffueraser: A diffusion model for video inpainting},
  author={Li, Xiaowen and Xue, Haolan and Ren, Peiran and Bo, Liefeng},
  journal={arXiv preprint arXiv:2501.10018},
  year={2025}
}

@inproceedings{fu2026effecterase,
  title={Effecterase: Joint video object removal and insertion for high-quality effect erasing},
  author={Fu, Yang and Zheng, Yike and Dai, Ziyun and Ding, Henghui},
  booktitle={Proceedings of the IEEE/CVF Conference on Computer Vision and Pattern Recognition},
  pages={2005--2014},
  year={2026}
}

@article{motamed2026void,
  title={Void: Video object and interaction deletion},
  author={Motamed, Saman and Harvey, William and Klein, Benjamin and Van Gool, Luc and Yuan, Zhuoning and Cheng, Ta-Ying},
  journal={arXiv preprint arXiv:2604.02296},
  year={2026}
}

@article{hu2026ideal,
  title={From Ideal to Real: Stable Video Object Removal under Imperfect Conditions},
  author={Hu, Jiagao and Chen, Yuxuan and Li, Fuhao and Wang, Zepeng and Wang, Fei and Zhou, Daiguo and Luan, Jian},
  journal={arXiv preprint arXiv:2603.09283},
  year={2026}
}

@article{kushwaha2026object,
  title={Object-wiper: Training-free object and associated effect removal in videos},
  author={Kushwaha, Saksham Singh and Nag, Sayan and Tian, Yapeng and Kulkarni, Kuldeep},
  journal={arXiv preprint arXiv:2601.06391},
  year={2026}
}

@article{zhang2025egolcd,
  title={EgoLCD: Egocentric Video Generation with Long Context Diffusion},
  author={Zhang, Liuzhou and Ye, Jiarui and Wang, Yuanlei and Zhong, Ming and Cao, Mingju and Xia, Wanke and Zeng, Bowen and Zhang, Zeyu and Tang, Hao},
  journal={arXiv preprint arXiv:2512.04515},
  year={2025}
}

@article{kong2024hunyuanvideo,
  title={Hunyuanvideo: A systematic framework for large video generative models},
  author={Kong, Weijie and Tian, Qi and Zhang, Zijian and Min, Rox and Dai, Zuozhuo and Zhou, Jin and Xiong, Jiangfeng and Li, Xin and Wu, Bo and Zhang, Jianwei and others},
  journal={arXiv preprint arXiv:2412.03603},
  year={2024}
}

@inproceedings{peebles2023scalable,
  title={Scalable diffusion models with transformers},
  author={Peebles, William and Xie, Saining},
  booktitle={Proceedings of the IEEE/CVF international conference on computer vision},
  pages={4195--4205},
  year={2023}
}

@inproceedings{li2022towards,
  title={Towards an end-to-end framework for flow-guided video inpainting},
  author={Li, Zhen and Lu, Cheng-Ze and Qin, Jianhua and Guo, Chun-Le and Cheng, Ming-Ming},
  booktitle={Proceedings of the IEEE/CVF conference on computer vision and pattern recognition},
  pages={17562--17571},
  year={2022}
}

@article{huang2023recent,
  title={Recent advances in artificial intelligence for video production system},
  author={Huang, YuFeng and Lv, ShiJuan and Tseng, Kuo-Kun and Tseng, Pin-Jen and Xie, Xin and Lin, Regina Fang-Ying},
  journal={Enterprise Information Systems},
  volume={17},
  number={11},
  pages={2246188},
  year={2023},
  publisher={Taylor \& Francis}
}


% 附录材料位置
% \input{supplementary.tex}

% \appendix
% % 下面几个是checklist里面涉及的
% \section{Datasets}
% \subsection{ROSE Benchmark}
% \subsection{VOR Benchmark}
% \subsection{EfffectLearner Benchmark}

% \section{Data Pre-processing Code}

% \section{Source Code}

\clearpage
\appendix

\section*{Technical Supplement of \\EffectLearner: World-Aware Object-Effect Reasoning for Real-World Video Object Removal}

\noindent\textbf{Table of Supplementary Materials}\\[6pt]
\begin{tabular*}{\columnwidth}{@{}l@{\extracolsep{\fill}}r@{}}
A \quad Benchmark Details & \pageref{sec:Benchmark-Details}\\
B \quad Training Data Details & \pageref{sec:Training-Data-Details}\\
C \quad EffectWorld Construction Details & \pageref{sec:EffectWorld-Construction}\\
D \quad Framework Implementation Details & \pageref{sec:Framework-Implementation}\\
E \quad More Evaluation on EffectWorld-Wild & \pageref{sec:EffectWorld-Wild-Evaluation}\\
F \quad Failure Cases and Analysis & \pageref{sec:failure_cases}\\
G \quad Ethical Statement & \pageref{sec:Ethical-Statement}
\end{tabular*}

% 讲述media supplement有静态网页
A static webpage archive for visual demonstrations is attached within the \textbf{Media Supplement}, which can be accessed by opening the contained \textbf{index.html} file after unzipping everything.

\section{Benchmark Details}
\label{sec:Benchmark-Details}
\subsection{ROSE-Bench}
ROSE~\cite{miao2026rose} has proposed ROSE-Bench to evaluate object-removal accuracy, effect-handling capability, and generalization to real-world scenes. 
Its \emph{synthetic paired benchmark} contains 60 paired source-mask-target video triplets, each evaluated over 49 frames. 
It covers the five object-induced effect categories, which are shadow, light source, reflection, mirror, and translucent effects. We use ROSE-Bench as the external paired benchmark for comparison with existing methods.

% \noindent\textbf{ROSE Training Dataset.}
% ROSE constructs a fully automated data-generation pipeline in Unreal Engine 5.3. It collects 28 high-quality base environments and divides them into 450 diverse indoor and outdoor scenes, covering urban areas, natural landscapes, artificial structures, vehicles, animals, and plants. For each scene, a target object is sampled and captured from different viewpoints, distances, and camera trajectories. Accurate frame-wise binary masks are generated using a customized post-processing shader, while views with severe occlusion or insufficient target visibility are filtered out. By toggling the target object's visibility while preserving identical scene configurations and camera motion, the pipeline produces spatially and temporally aligned triplets of source videos, object-removed target videos, and mask videos. The resulting dataset contains 16,678 video triplets, each comprising 90 frames over six seconds at a resolution of $1920\times1080$. It covers six categories: common objects with minimal environmental interactions, light sources, mirrors, reflections, shadows, and translucent effects. Diverse weather, illumination, and object configurations further improve data diversity~\cite{miao2026rose}.
\subsection{EffectWorld-Eval}
\ourbencheval{} is a paired benchmark containing 33 complex-effect videos, each with a source video, an object mask, and an object-removed target. Its composition follows the main challenges studied in this work: 29 videos contain compositional effects, 8 exhibit spatially detached or weakly correlated object-effect relations, and all 33 involve object motion. We additionally retain the five effect labels adopted by ROSE for category-level analysis. Since these labels are non-exclusive, the benchmark contains 32 shadow, 10 mirror, 10 reflection, 6 light, and 6 translucent cases.

\subsection{EffectWorld-Wild}
\ourbenchwild{} evaluates object removal in open-world videos without paired targets. It contains 70 source-mask pairs, comprising 59 curated real-world videos and 11 videos from DAVIS. Each sample provides a manually verified target-object mask. The absence of object-removed targets makes this benchmark suitable for no-reference and subjective evaluation.

The 59 non-DAVIS videos were collected from Pexels and manually screened for clearly identifiable removal targets, sufficient temporal visibility, and observable interactions between the target and its surroundings. The verified diagnostic annotations include 12 physical-trace cases and six fast-motion cases. We also retain the five non-exclusive effect labels adopted by ROSE, resulting in 41 shadow, 11 mirror, eight reflection, seven light, and five translucent cases.

We obtain target masks through a semi-automatic annotation procedure. A zero-shot detector first localizes the target using a category-specific text prompt, after which SAM2 propagates the selected box through the video. We manually inspect the mask overlays and multi-frame contact sheets for every sample. Tracking errors, missing object parts, and background leakage are corrected by rerunning SAM2 with manually specified box prompts.

\section{Training Data Details}
\label{sec:Training-Data-Details}
The complete \datasetname{} UE collection contains 11,092 valid paired triplets. It includes complex-effect examples that complement the standard effect supervision provided by ROSE. The complete collection, however, is not exhaustively annotated along the compositional, weak-correlation, long-tail, and dynamic dimensions.The formal training manifest uses 11,048 \datasetname{} triplets together with 16,663 ROSE triplets, yielding 27,711 training samples.
\begin{figure*}[t]
\centering
\includegraphics[width=\textwidth]{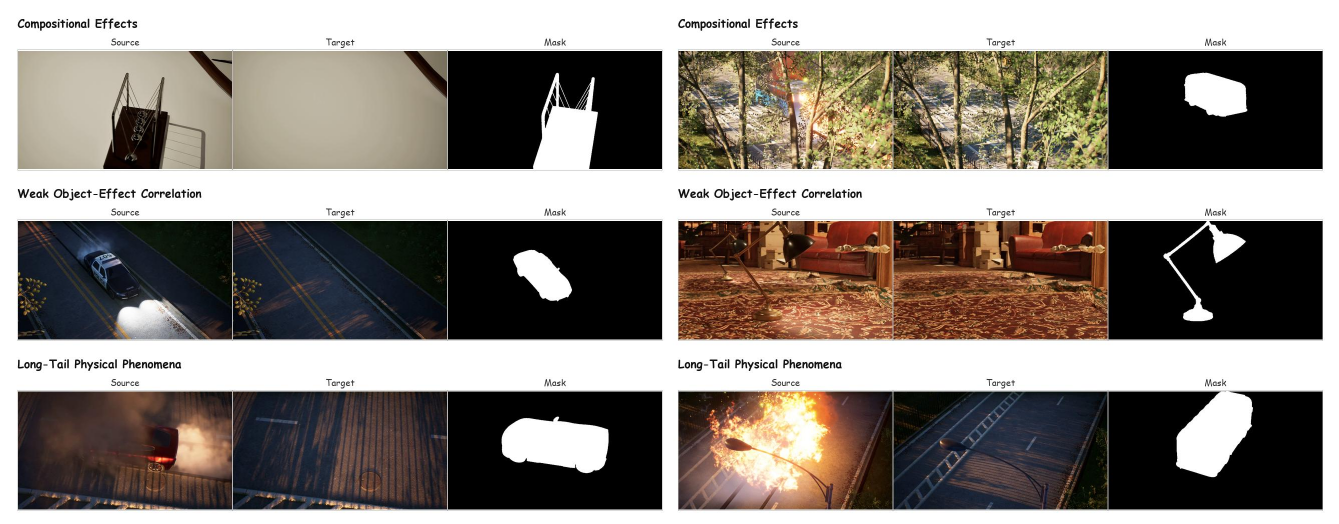}
\renewcommand{\thefigure}{\Roman{figure}}
\caption{Representative paired triplets from \textbf{\datasetname{}}. Each example shows the source video, object-removed target, and target-object mask. The examples cover compositional effects, long-tail physical phenomena, and weak object--effect correlations.}
\label{fig:effectworld_examples}
\end{figure*}

\section{EffectWorld Construction Details}
\label{sec:EffectWorld-Construction}

\boldparagraph{UE triplet generation.}
\begin{algorithm}[htbp]

\caption{UE paired-triplet generation}
\label{alg:ue_triplet_generation}
\begin{algorithmic}[1]
\REQUIRE UE scene $\mathcal{S}$; actor tag $\tau$; iterations $N$; sequence length $T$
\ENSURE Aligned source--target--mask triplets $\mathcal{D}$
\STATE $\mathcal{A}\leftarrow\{a\in\mathcal{S}\mid a\text{ has tag }\tau\}$
\STATE Create or load a white unlit emissive material $m_{\mathrm{white}}$
\FOR{each target actor $a\in\mathcal{A}$}
    \STATE Save the original transform of $a$
    \FOR{$n=1$ to $N$}
        \STATE Spawn a CineCameraActor and set its tracking focus to $a$
        \STATE Sample motion intensities and the end transform of $a$
        \STATE Sample start/end camera poses around the moving target
        \STATE Create a 15-fps LevelSequence $L$ of length $T$
        \STATE Add target/camera transform tracks and a camera-cut track to $L$
        \STATE $\mathcal{G}\leftarrow\{a\}\cup\operatorname{AttachedActors}(a)$
        \STATE Show $\mathcal{G}$ and render $L$ to \texttt{01\_Source}
        \STATE Hide $\mathcal{G}$ and render $L$ to \texttt{02\_Target}
        \STATE Hide non-target actors; show $\mathcal{G}$; save its materials
        \STATE Replace target materials with $m_{\mathrm{white}}$ and render $L$ to \texttt{03\_Mask}
        \STATE Restore all hidden actors and original target materials
        \STATE Add the three rendered streams to $\mathcal{D}$
    \ENDFOR
\ENDFOR
\RETURN $\mathcal{D}$
\end{algorithmic}
\end{algorithm}

The generator identifies target actors with the tag \texttt{rose} and creates 20 randomized sequences per target. It first records the original target transform. For each sequence, a translation intensity $s_m\sim\mathcal{U}(0.1,1.0)$ scales a maximum displacement of 150 Unreal units, after which the $x$ and $y$ displacements are independently sampled from $\mathcal{U}(-150s_m,150s_m)$ while the vertical position remains fixed. A rotation intensity $s_r\sim\mathcal{U}(0.1,1.0)$ similarly scales a maximum yaw displacement of $360^\circ$. Roll, pitch, and scale remain unchanged.

Each iteration creates an Unreal Engine Cine Camera Actor with tracking focus on the target. Let $b$ be the length of the target bounding-box extent vector. The base camera distance is $\max(3.5b,50)$. The initial azimuth is sampled from $\mathcal{U}(0,2\pi)$ and changes by at most 0.2 radians; the initial polar angle is sampled from $\mathcal{U}(0.7,1.0)$ and changes by at most 0.1 radians. The initial distance is sampled from $[1.0,1.3]$ times the base distance, and its endpoint is sampled from $[0.9,1.1]$ times the initial distance. Additional yaw and pitch perturbations are sampled from $[-18^\circ,18^\circ]$ and $[-12^\circ,12^\circ]$.

The start and end transforms of the camera and target are written into a shared LevelSequence with separate 3D transform tracks and a camera-cut track. Its display rate is 15 fps. In the supplied generator configuration, \texttt{TOTAL\_FRAMES} is 150 and the tracks span frames 0 to 150.

The same LevelSequence is rendered sequentially into \texttt{01\_Source}, \texttt{02\_Target}, and \texttt{03\_Mask}.  In the mask pass, all other visible scene actors are temporarily hidden and each target mesh material is replaced with an automatically created white unlit emissive material. Movie Render Queue uses the deferred rendering pass and PNG image-sequence output. Frames are named by frame number and stored under \texttt{<actor>/Iter\_<id>/<pass>}.

\boldparagraph{Standardization and validation.}
Let an input frame have width $W$ and height $H$. We compute
\begin{equation}
s=\max\left(\frac{1280}{W},\frac{704}{H}\right),
\end{equation}
resize the frame to $(\operatorname{round}(sW),\operatorname{round}(sH))$, and apply a center crop to $1280\times704$. Source and target frames are converted to RGB, resized using bicubic interpolation, and saved as JPEG images with quality 95. Masks are converted to grayscale, resized using nearest-neighbor interpolation, binarized using the rule $v>127$, and saved as PNG images.

The standardized directories retain the names \texttt{01\_Source}, \texttt{02\_Target}, and \texttt{03\_Mask}. Output frames use six-digit indices beginning with \texttt{000000}; source and target frames use the extension \texttt{.jpg}, while masks use \texttt{.png}.

Each manifest row contains \texttt{src\_video}, \texttt{tgt\_video}, \texttt{mask\_video}, \texttt{prompt}, \texttt{dataset}, \texttt{sample\_id}, and \texttt{frames}. The complete standardized dataset contains 11,092 valid triplets, including 3,428 sequences with 90 frames and 7,664 with 150 frames.

\section{Framework Implementation Details}
\label{sec:Framework-Implementation}
%%%%%%%%%%%%%%%5 不要动
\subsection{Structured Effect-Analysis Prompt.}
The complete \textbf{Structured Effect-Analysis Prompt} is provided in Figure~\ref{fig:prompt}, including the analysis role prompt, the general effect-analysis instruction prompt, and the sample-level removal instruction prompt.

\begin{figure*}[t]
\centering
\includegraphics[width=0.7\linewidth]{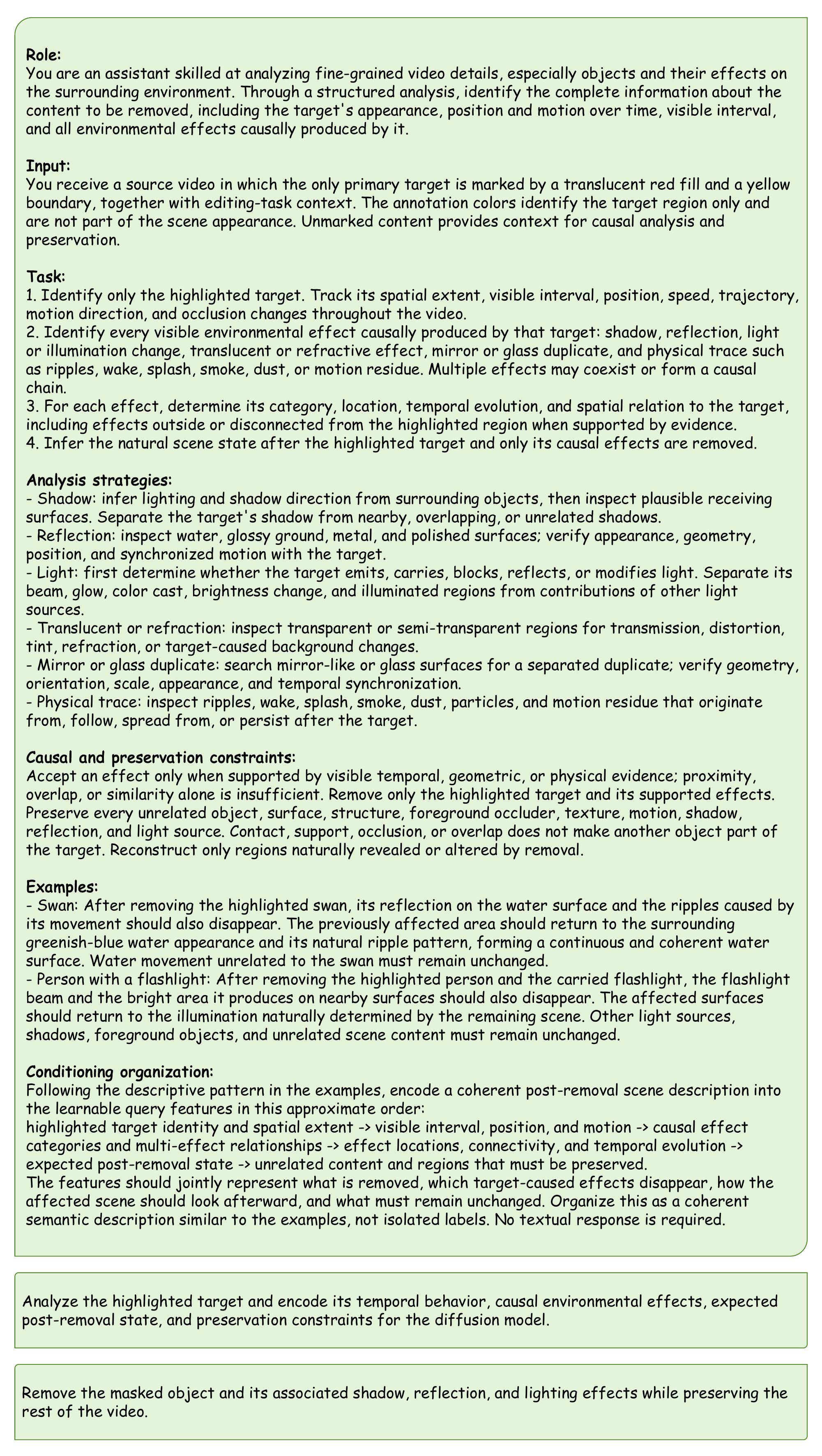}
\renewcommand{\thefigure}{\Roman{figure}}
\caption{The complete \textbf{Structured Effect-Analysis Prompt}. From top to bottom are the analysis role prompt, the general effect-analysis instruction prompt, and the sample-level removal instruction prompt.}
\label{fig:prompt}
\end{figure*}

\subsection{Mask Construction and Objectives}
Frame-level object masks are first resized to the latent spatial resolution using nearest-neighbor interpolation. The source timeline is divided into $T_{\mathrm{latent}}$ windows by $\operatorname{linspace}(0,T_{\mathrm{src}},T_{\mathrm{latent}}+1)$, and masks within each window are merged by element-wise maximum. The resulting four-channel $M_{\mathrm{union}}$ is used as the DiT condition. Applying one $3\times3$ max-pooling dilation produces $M_{\mathrm{loss}}$ for loss weighting.

The weighted flow-matching objective uses
\begin{equation}
W=4M_{\mathrm{loss}}+2(1-M_{\mathrm{loss}})
\end{equation}
to regress $v^*=\epsilon-x_0$. The effect region $E$ is obtained by thresholding the channel-averaged latent difference $\lvert x_0-x_s\rvert$ at 0.06, applying one spatial dilation, and taking its maximum with $M_{\mathrm{loss}}$. The Semantic Effect Loss is
\begin{equation}
\begin{aligned}
\mathcal{L}_{\mathrm{semantic}}
={}&0.30\,\operatorname{Mean}_{E}
\|\hat{x}_0-x_0\|_2^2 \\
&+0.20\,\operatorname{Mean}_{1-E}
\|\hat{x}_0-x_s\|_2^2.
\end{aligned}
\end{equation}

For temporal supervision, centroid displacement between adjacent masks provides $(d_x,d_y)$. The negative displacement is broadcast over the latent grid to form a translational pseudo-flow compatible with backward sampling. The Motion Consistent Loss compares motion-compensated temporal residuals of the predicted clean latent and target clean latent inside the union of the dilated effect region and adjacent object masks, with weight 0.20. The difference-mask-predictor branch is present in the repository but disabled in formal training because its loss weight remains zero. The resulting loss is finally multiplied by the scheduler-provided timestep weight, whose 1,000 entries are normalized to sum to 1,000 and emphasize middle timesteps.

\subsection{Target-Highlighted Video Input}
Algorithm~\ref{alg:target_highlight} details the construction of the target-highlighted video. 

\begin{algorithm}[htbp]

\caption{Target-Highlighted Video Input}
\label{alg:target_highlight}
\begin{algorithmic}[1]
\REQUIRE Source video $\mathbf{V}_{\mathrm{src}}\in\mathbb{R}^{T\times H\times W\times3}$; binary object mask $\mathbf{M}\in\{0,1\}^{T\times H\times W}$; highlight color $\mathbf{c}_{\mathrm{hl}}=(255,0,0)$; boundary color $\mathbf{c}_{\mathrm{bd}}=(255,255,0)$; blending coefficient $\alpha=0.15$
\ENSURE Target-highlighted video $\mathbf{V}_{\mathrm{hl}}$
\STATE Broadcast $\mathbf{M}$ over the three color channels
\STATE $\mathbf{V}_{\mathrm{hl}}\leftarrow(1-\alpha\mathbf{M})\odot\mathbf{V}_{\mathrm{src}}+\alpha\mathbf{M}\odot\mathbf{c}_{\mathrm{hl}}$
\STATE $w\leftarrow\max\!\left(2,\operatorname{round}(\min(H,W)/176)\right)$
\STATE $\mathbf{B}\leftarrow\operatorname{Boundary}(\mathbf{M},w)$
\STATE Set $\mathbf{V}_{\mathrm{hl}}$ to $\mathbf{c}_{\mathrm{bd}}$ at boundary pixels $\mathbf{B}$
\RETURN $\mathbf{V}_{\mathrm{hl}}$
\end{algorithmic}
\end{algorithm}

% \begin{algorithm}[t]
% \caption{Structured Effect-Analysis Prompt}
% \label{alg:effect_prompt}
% \begin{algorithmic}[1]
% \REQUIRE Target-highlighted video $\mathbf{V}_{\mathrm{hl}}$; analysis-role prompt $P_{\mathrm{sys}}$; general effect-analysis instruction $P_{\mathrm{inst}}$; sample-level removal instruction $P_{\mathrm{task}}$; vision-language model $\mathrm{VLM}$
% \ENSURE Removal-relevant VLM hidden states
% \STATE Specify the VLM's object-effect analysis role in $P_{\mathrm{sys}}$
% \STATE Specify target grounding, induced-effect analysis, post-removal inference, and preservation constraints in $P_{\mathrm{inst}}$
% \STATE Read the sample-level removal objective from $P_{\mathrm{task}}$
% \STATE $P\leftarrow[P_{\mathrm{sys}};P_{\mathrm{inst}};P_{\mathrm{task}}]$
% \STATE Apply $\mathrm{VLM}(\mathbf{V}_{\mathrm{hl}},P)$ to obtain removal-relevant hidden states
% \RETURN Removal-relevant VLM hidden states
% \end{algorithmic}
% \end{algorithm}

\subsection{Motion-Aware Temporal Mask Union}
Algorithm~\ref{alg:temporal_mask_union} describes how frame-level object masks are aligned with the latent timeline. 

\begin{algorithm}[t]
\caption{Motion-Aware Temporal Mask Union}
\label{alg:temporal_mask_union}
\begin{algorithmic}[1]
\REQUIRE Frame-level masks $M=\{M^{(i)}\}_{i=1}^{T}$; $T_{\mathrm{latent}}$ latent temporal positions; latent spatial size $(H_l,W_l)$
\ENSURE DiT mask condition $M_{\mathrm{union}}$; supervision mask $M_{\mathrm{loss}}$
\FOR{$i=1$ to $T$}
    \STATE $\widetilde{M}^{(i)}\leftarrow\operatorname{Resize}_{\mathrm{nearest}}(M^{(i)},H_l,W_l)$
\ENDFOR
\STATE Divide the $T$ frames into $T_{\mathrm{latent}}$ windows $\{\Omega_k\}_{k=1}^{T_{\mathrm{latent}}}$ using $\operatorname{linspace}(0,T,T_{\mathrm{latent}}+1)$
\FOR{$k=1$ to $T_{\mathrm{latent}}$}
    \STATE $M_{\mathrm{union}}^{k}\leftarrow\max_{i\in\Omega_k}\widetilde{M}^{(i)}$ \hfill (element-wise)
    \STATE Replicate $M_{\mathrm{union}}^{k}$ over four channels for DiT conditioning
    \STATE $M_{\mathrm{loss}}^{k}\leftarrow\operatorname{MaxPool}_{3\times3}(M_{\mathrm{union}}^{k})$
\ENDFOR
\RETURN $M_{\mathrm{union}},M_{\mathrm{loss}}$
\end{algorithmic}
\end{algorithm}

\subsection{Motion-Aware Temporal Mask Union}
Algorithm~\ref{alg:motion_consistent_loss} presents the Motion Consistent Loss. 
\begin{algorithm}[htbp]
\caption{Motion Consistent Loss}
\label{alg:motion_consistent_loss}
\begin{algorithmic}[1]
\REQUIRE Predicted clean latents $\{\hat{x}_0^k\}_{k=1}^{K}$; target clean latents $\{x_0^k\}_{k=1}^{K}$; loss masks $\{M_{\mathrm{loss}}^k\}_{k=1}^{K}$; temporal supervision regions $\{R_k\}_{k=2}^{K}$; backward warping operator $\mathcal{W}$ using bilinear interpolation, border padding, and aligned corners; numerical constant $\varepsilon$
\ENSURE Motion consistent loss $\mathcal{L}_{\mathrm{motion}}$
\STATE $N\leftarrow0$; $D\leftarrow0$
\FOR{$k=2$ to $K$}
    \STATE $c_{k-1}\leftarrow\dfrac{\sum_{p}p\,M_{\mathrm{loss}}^{k-1}(p)}{\sum_{p}M_{\mathrm{loss}}^{k-1}(p)+\varepsilon}$
    \STATE $c_k\leftarrow\dfrac{\sum_{p}p\,M_{\mathrm{loss}}^k(p)}{\sum_{p}M_{\mathrm{loss}}^k(p)+\varepsilon}$
    \STATE $d_k\leftarrow c_k-c_{k-1}$
    \STATE Broadcast $-d_k$ over the latent grid to form the backward pseudo-flow $f_k$
    \STATE $\hat{\Delta}_k\leftarrow\hat{x}_0^k-\mathcal{W}(\hat{x}_0^{k-1},f_k)$
    \STATE $\Delta_k\leftarrow x_0^k-\mathcal{W}(x_0^{k-1},f_k)$
    \STATE $N\leftarrow N+\sum_{p}R_k(p)\|\hat{\Delta}_k(p)-\Delta_k(p)\|_2^2$
    \STATE $D\leftarrow D+\sum_{p}R_k(p)$
\ENDFOR
\STATE $\mathcal{L}_{\mathrm{motion}}\leftarrow N/(D+\varepsilon)$
\RETURN $\mathcal{L}_{\mathrm{motion}}$
\end{algorithmic}
\end{algorithm}

\begin{table*}[t]
\centering
\small
\setlength{\tabcolsep}{4pt}
\renewcommand{\thetable}{\Roman{table}}
\begin{tabular}{lccccc}
\toprule
Method
& Target Removal
& Effect Removal
& Background Preservation
& Temporal Rendering
& Overall \\
\midrule
ProPainter~\cite{zhou2023propainter}
& 2.696 & 1.842 & 2.152 & 2.114 & 2.201 \\
DiffuEraser~\cite{li2025diffueraser}
& 2.860 & 2.117 & 2.529 & 2.526 & 2.508 \\
VACE~\cite{jiang2025vace}
& 1.237 & 1.149 & 1.719 & 1.921 & 1.507 \\
ROSE~\cite{miao2026rose}
& 2.769 & 2.287 & 2.202 & 2.152 & 2.352 \\
EffectErase~\cite{fu2026effecterase}
& \underline{3.637} & \underline{3.231} & \textbf{3.120} & \underline{3.094} & \underline{3.270} \\
\methodname{} (Ours)
& \textbf{3.778} & \textbf{3.547} & \underline{3.076} & \textbf{3.155} & \textbf{3.389} \\
\bottomrule
\end{tabular}
\caption{Results of various methods with the LLM Judge in \ourbenchwild. The best results are in \textbf{bold}, while the second-best results are \underline{underlined}. }
\label{tab:wild_judge}
\end{table*}

\begin{table*}[t]
\centering
\small
\setlength{\tabcolsep}{4pt}
\renewcommand{\thetable}{\Roman{table}}
\begin{tabular}{lccccc}
\toprule
Method & Target Removal & Effect Removal & Background Preservation & Temporal Rendering & Overall \\
\midrule
ProPainter~\cite{zhou2023propainter}
& \underline{2.950} & 1.850 & 1.950 & 1.050 & 1.950 \\

DiffuEraser~\cite{li2025diffueraser}
& \underline{2.950} & 1.950 & 1.100 & \underline{2.100} & 2.025 \\

VACE~\cite{jiang2025vace}
& 1.000 & 1.000 & 1.000 & 1.150 & 1.038 \\

ROSE~\cite{miao2026rose}
& 3.000 & 2.050 & 1.850 & 1.900 & 2.200 \\

EffectErase~\cite{fu2026effecterase}
& \textbf{3.900} & \underline{3.000} & \underline{3.000} & \textbf{2.950} & \underline{3.213} \\

\methodname{} (Ours)
& \textbf{3.900} & \textbf{3.900} & \textbf{3.950} & \textbf{2.950} & \textbf{3.675} \\
\bottomrule
\end{tabular}
\caption{Results of various methods with human evaluation on \ourbenchwild. The best results are in \textbf{bold}, while the second-best results are \underline{underlined}.}
\label{tab:wild_human}
\end{table*}

\section{More Evaluation on EffectWorld-Wild}
\label{sec:EffectWorld-Wild-Evaluation}

\subsection{Fine-grained Object Removal Evaluation Protocol}

We introduce a unified fine-grained evaluation protocol for both LLM-based and human assessment. 
Given the source video and anonymized candidate results, evaluators inspect the complete temporal sequence and independently score each result along four dimensions. 
\textit{Target Removal} measures whether the highlighted object is completely removed in all frames without visible remnants. 
\textit{Effect Removal} evaluates whether its causal effects, such as shadows, reflections, illumination changes, ripples, smoke, and motion traces, are also eliminated, including effects extending beyond the object mask. 
\textit{Background Preservation} assesses whether unrelated objects, structures, textures, and scene regions remain unchanged, thereby penalizing excessive or unintended edits. 
\textit{Temporal Rendering} measures the naturalness of the reconstructed content, the absence of visual artifacts, and its consistency across frames. 
Each dimension is rated on an integer scale from 1 (substantial failure) to 4 (complete and nearly artifact-free removal). 
To avoid favoring visually appealing but incomplete edits, rendering quality is not rewarded when the target or its induced effects remain. 
The \textit{Overall} score is computed as the equally weighted average of the four dimensions, providing a unified measure of removal completeness, edit exclusivity, and temporal restoration quality.

\begin{figure*}[t]
\centering
\includegraphics[width=0.95\linewidth]{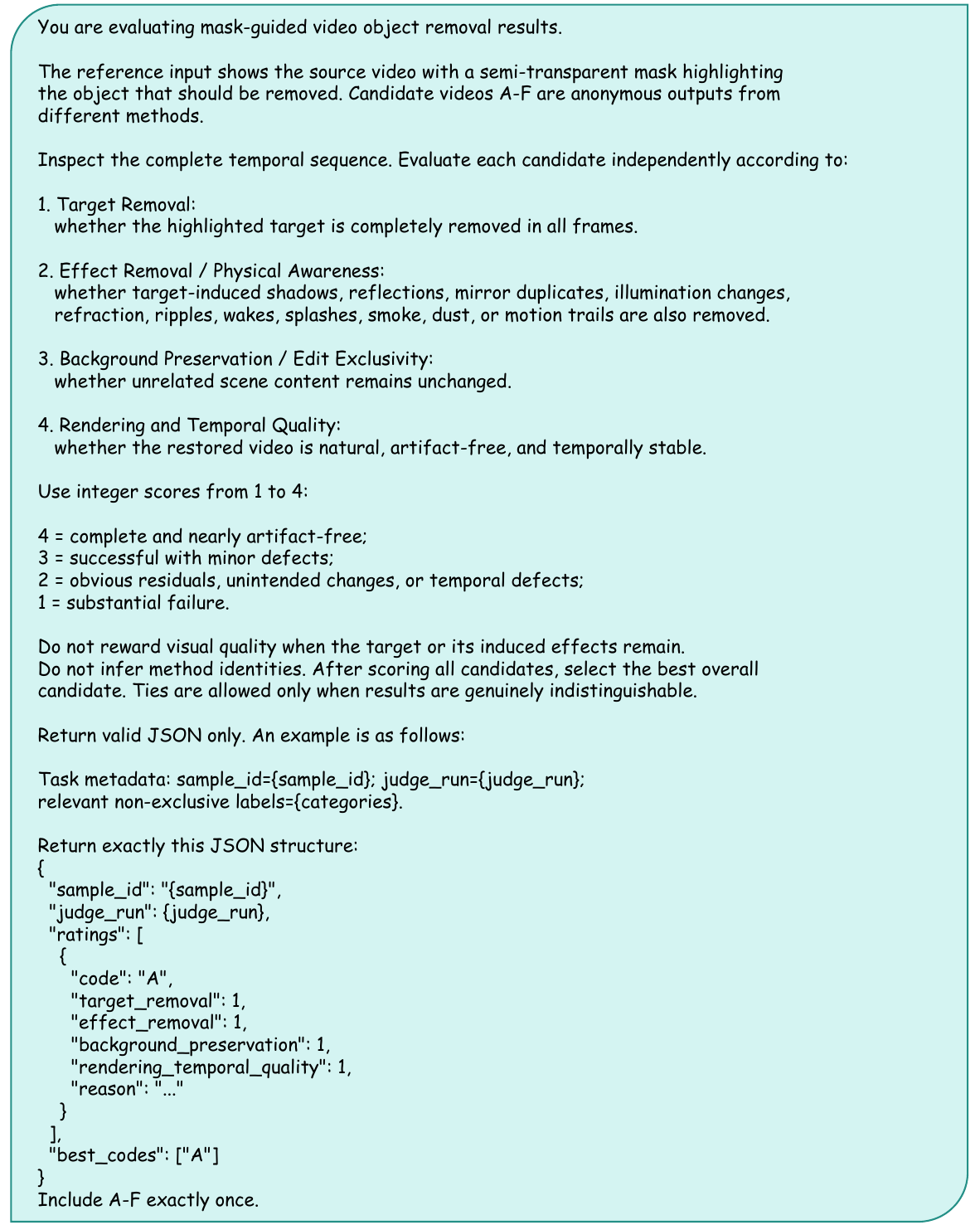}
\renewcommand{\thefigure}{\Roman{figure}}
\caption{The complete prompt for LLM judge in the fine-grained object removal evaluation protocol.}
\label{fig:prompt}
\end{figure*}

\subsection{LLM-as-a-Judge Evaluation}
% Pearson Result: 0.9837
We apply GPT-5.4~\cite{openai2026gpt54} as the LLM judge. Table~\ref{tab:wild_judge} shows that \methodname{} achieves the highest overall LLM-as-a-Judge score, ranking first in target removal, effect removal, and temporal rendering. Its pronounced advantage in effect removal demonstrates a stronger capability to recognize and eliminate target-induced effects beyond the annotated object region. EffectErase is the closest competitor and slightly outperforms our method in background preservation, but its lower removal and temporal scores indicate a less favorable balance between complete erasure and coherent restoration. Conventional inpainting methods and ROSE retain moderate visual quality but remain limited in handling complex causal effects, while VACE performs poorly across all dimensions. 

%%%%%%%%%%% 之前自己提出的llm-judge的表

\subsection{Human Evaluation}
We recruit 20 human volunteers to evaluate the videos generated by each method on \ourbenchwild{} following the proposed protocol.
Each volunteer assigned an integer score from 1 to 4 to every evaluation dimension, and the resulting ratings were averaged to obtain the human-evaluation score for each method. 
Table~\ref{tab:wild_human} shows that \methodname{} achieves the highest overall human-evaluation score, ranking first in effect removal and background preservation while tying for the best performance in target removal and temporal rendering. Compared with EffectErase, the closest competitor, our method achieves substantially better effect removal and background preservation, indicating more complete removal of target-induced effects without damaging unrelated scene content. Other methods exhibit clear trade-offs between removal accuracy, preservation, and temporal quality, resulting in lower overall scores. \textbf{Notably, the overall scores from human and LLM-based evaluations exhibit a strong Pearson correlation of $r=0.9837$, demonstrating high agreement between the LLM judge and human}. This consistency suggests that LLM-as-a-Judge can serve as an efficient and reliable proxy for rapid, high-quality evaluation in practical settings.

\section{Failure Cases and Analysis}
\label{sec:failure_cases}
\begin{figure*}[htbp]
    \centering
    \renewcommand{\thefigure}{\Roman{figure}}
    \begin{minipage}[t]{0.8\textwidth}
        \centering
        \includegraphics[
            width=\linewidth
        ]{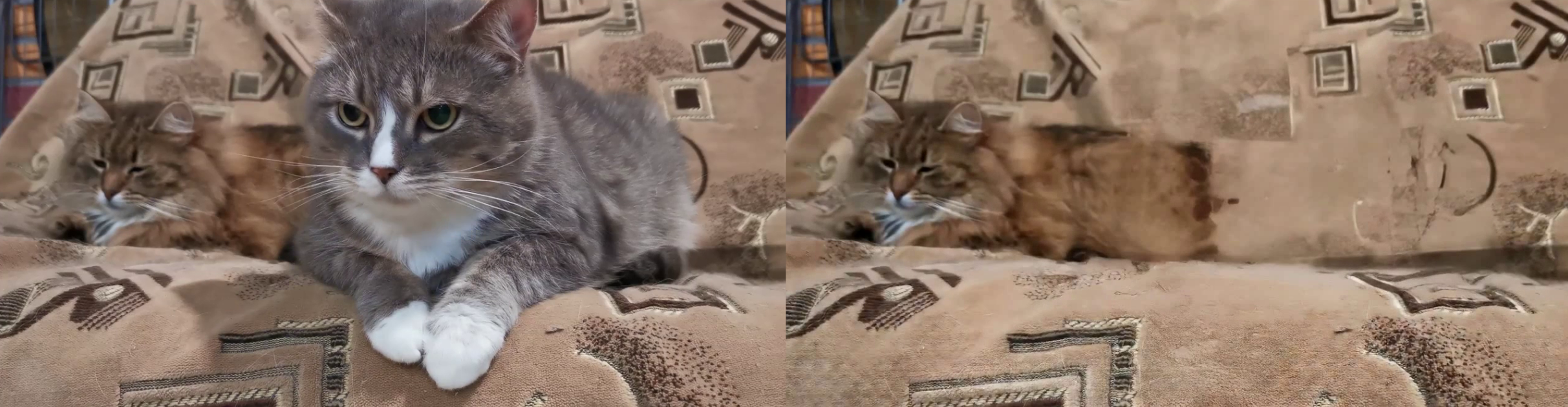}
        \textbf{(a) Large occlusion and complex background texture}
    \end{minipage}
    
    \vspace{1em} % 两张子图垂直间距，按需调整
    \begin{minipage}[t]{0.8\textwidth}
        \centering
        \includegraphics[
            width=\linewidth
        ]{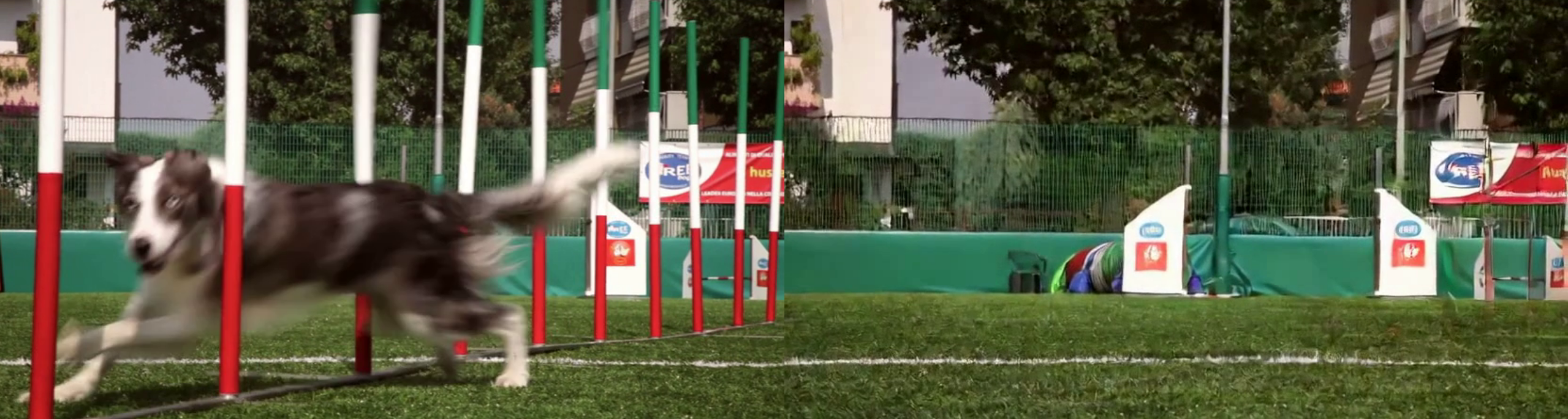}
        \textbf{(b) Complex motion and object occlusion}
    \end{minipage}
    \caption{
    Representative failure cases of \methodname{}.
    Each example compares the original input with our removal result.
    Large foreground occlusion may lead to over-smoothed or distorted background reconstruction, while fast motion and repeated occlusion by thin structures may introduce local structural artifacts.
    }
    \label{fig:failure_cases}
\end{figure*}

Our model still has some room for improvement in some complex scenarios. As shown in Figure~\ref{fig:failure_cases}, two failure modes are observed. 
1) In the cat example, the target occupies a large foreground region and occludes both a highly textured sofa and part of another cat. Although the target is removed, the reconstructed region contains over-smoothed textures and local distortions near the remaining cat. 
2) In the dog example, severe motion blur and repeated occlusion cause the model to incorrectly remove parts of the thin poles that should have been preserved.
These cases indicate that recovering high-frequency content occluded by large targets and maintaining precise alignment under complex motion or occlusion remain challenging.

\section{Ethical Statement}
\label{sec:Ethical-Statement}
Our self-constructed \datasetname{} training data are rendered in Unreal Engine and contain no captured footage or biometric information from real individuals, and public ROSE data are used under their original terms. 
DAVIS and publicly accessible, free-to-use Pexels videos are used only for evaluation under their respective licenses, without collecting identity or biometric annotations. 
Although video object removal supports benign applications such as editing and post-production, it may be misused to conceal evidence or fabricate misleading content. 
We therefore plan to release our method for research purposes with responsible-use terms prohibiting deepfakes, disinformation, impersonation, and deceptive alteration, while encouraging disclosure of edited media and compliance with applicable privacy, copyright, and data-protection requirements.

\end{document}